\documentclass[10pt,twocolumn,letterpaper]{article}

\usepackage{iccv}              
\usepackage[accsupp]{axessibility}  

\definecolor{iccvblue}{rgb}{0.21,0.49,0.74}
\definecolor{goldenrod}{RGB}{218, 165, 32}
\usepackage[pagebackref,breaklinks,colorlinks,allcolors=iccvblue]{hyperref}

\def\paperID{8066} 
\def\confName{ICCV}
\def\confYear{2025}

\title{Towards Robust Defense against Customization via Protective Perturbation Resistant to Diffusion-based Purification}

\author{
    Wenkui Yang$^{1,2}$, Jie Cao$^{1}$, Junxian Duan$^{1}$, Ran He$^{1,2}$\thanks{Corresponding author.}\\
    {$^{1}$MAIS \& NLPR, Institute of Automation, Chinese Academy of Sciences}\\
    {$^{2}$School of Artificial Intelligence, University of Chinese Academy of Sciences}\\
    {\small \texttt{yangwenkui.03@gmail.com,}}
    {\small \texttt{\{jie.cao,junxian.duan\}@cripac.ia.ac.cn,}}
    {\small \texttt{rhe@nlpr.ia.ac.cn}}
}

\usepackage{multirow}
\begin{document}
\maketitle
\begin{abstract}

Diffusion models like Stable Diffusion have become prominent in visual synthesis tasks due to their powerful customization capabilities, which also introduce significant security risks, including deepfakes and copyright infringement. 
In response, a class of methods known as protective perturbation emerged, which mitigates image misuse by injecting imperceptible adversarial noise.
However, purification can remove protective perturbations, thereby exposing images again to the risk of malicious forgery.

In this work, we formalize the anti-purification task, highlighting challenges that hinder existing approaches, and propose a simple diagnostic protective perturbation named \textbf{AntiPure}.
AntiPure exposes vulnerabilities of purification within the ``purification-customization'' workflow, owing to two guidance mechanisms: 1) Patch-wise Frequency Guidance, which reduces the model’s influence over high-frequency components in the purified image, and 2) Erroneous Timestep Guidance, which disrupts the model’s denoising strategy across different timesteps.
With additional guidance, AntiPure embeds imperceptible perturbations that persist under representative purification settings, achieving effective post-customization distortion. 
Experiments show that, as a stress test for purification, AntiPure achieves minimal perceptual discrepancy and maximal distortion, outperforming other protective perturbation methods within the purification-customization workflow.

\end{abstract}    
\section{Introduction}
\label{sec:intro}

\begin{figure}[t]
\centering
\includegraphics[width=0.9\linewidth]{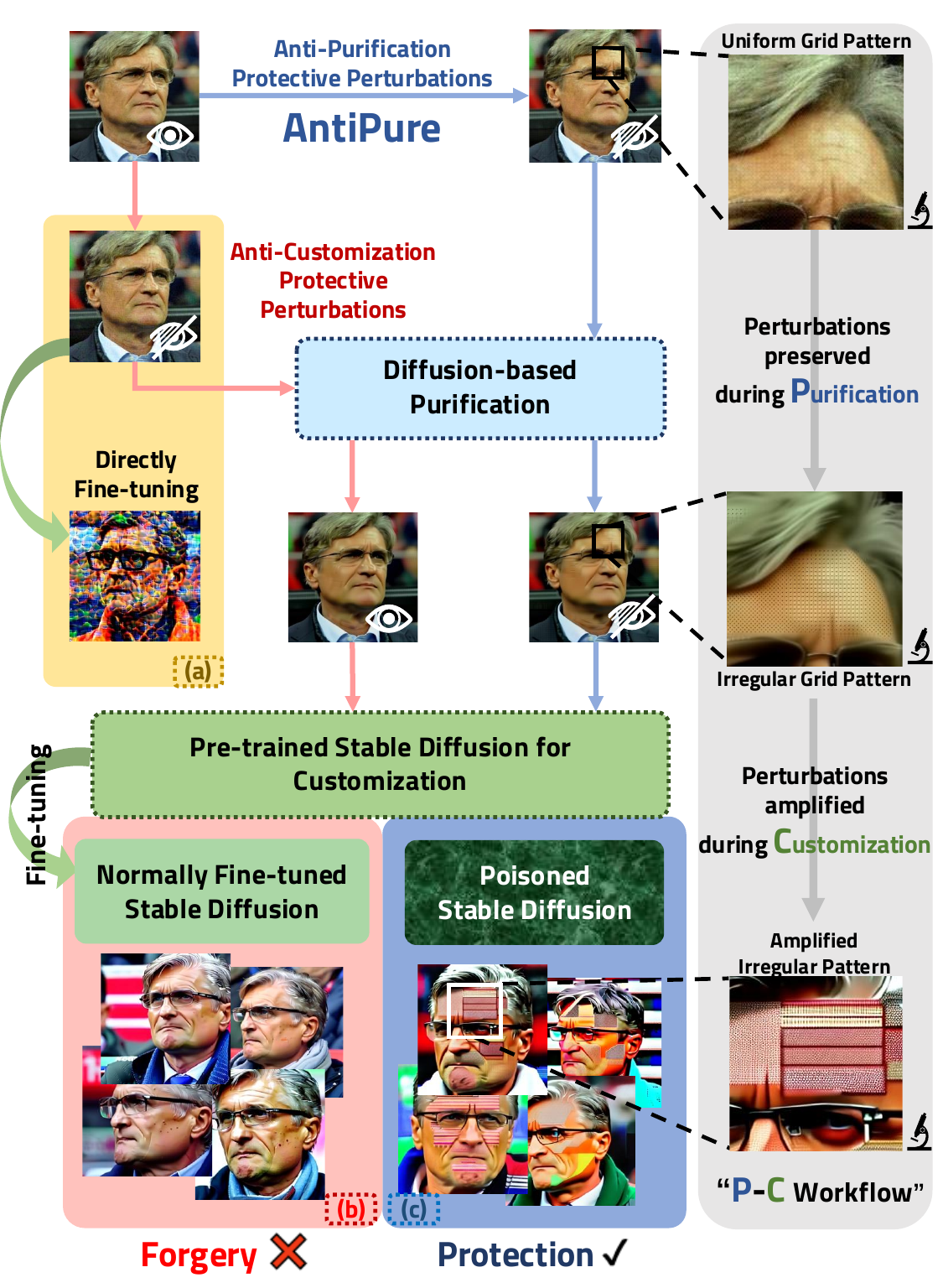}
\vspace{-5pt}
\caption{Overview diagram of ``Purification-Customization" workflow. \textcolor{goldenrod}{a).} Protective perturbations use small noises to distort the outputs of fine-tuned diffusion models. \textcolor{red}{b).} However, existing methods can be removed by diffusion-based purification. \textcolor{blue}{c).} We propose a simple diagnostic method called \textbf{AntiPure}, which achieves protective perturbations resistant to purification and makes customization outputs more distinguishable.}
\vspace{-15pt}
\label{fig:workflow}
\end{figure}

Since the landmark developments of Denoising Diffusion Probabilistic Models (DDPMs) \cite{sohl2015deep,ho2020denoising} and Latent Diffusion Models (LDMs) \cite{rombach2022high}, Diffusion Models (DMs) have virtually dominated every subtask of visual generation. 
This triumph can be attributed to the readily available open-source Stable Diffusion (SD) \cite{stable_diffusion}, which enables numerous fine-tuning and editing techniques for user-friendly customization. 
However, these advances also pose risks, including the proliferation of deepfakes and infringements of portrait rights and intellectual property.

In response, recent studies \cite{liang2023adversarial,liang2023mist,van2023anti,wang2024simac,xu2024perturbing,liu2024metacloak,ahn2025nearly,song2025idprotector} adapt adversarial attacks \cite{goodfellow2014explaining,kurakin2016adversarial,madry2017towards} to DMs, creating ``poisoned" samples that impede concept comprehension during fine-tuning \textbf{Customization}. 
Specifically, these \textbf{Anti-Customization} approaches—collectively termed \textbf{Protective Perturbations}—employ white-box attacks to find the most adversarial perturbation noises within the \( l_\infty \)-box, thereby distorting DMs' outputs.

Unfortunately, these adversarial protective perturbations can be nullified by diffusion-based \textbf{Purification} \cite{nie2022diffusion,xiao2023densepure,zhao2024can}.
Such methods diffuse adversarial samples to a fixed timestep and then remove both adversarial and diffusion noises during denoising.
As shown in \cref{fig:workflow}, prior methods rarely consider how to preserve protective effects when purification precedes fine-tuning. This two-stage Purification-Customization (P-C) workflow therefore renders existing protective perturbations vulnerable and largely ineffective.

In this paper, we show the possibility of purification-resistant protective perturbations in representative P-C settings, despite the strong denoising capability of diffusion-based purification.
First, we formalize how to achieve that within the P-C workflow.
We observe that diffusion models—as a class of probabilistic models—produce outputs that can become highly unpredictable at the fine scale required by adversarial attacks, thereby diminishing the effectiveness of adaptive attacks.
Based on this, rather than attempting to preserve the anti-customization perturbations during purification, we propose \textbf{anti-purification} perturbations that directly target the purification model to probe and expose its weaknesses.
Through experiments, we analyze the differences between anti-customization and anti-purification, identifying three core characteristics of purification models that make anti-purification more challenging: 1) lack of vulnerable network components, 2) training-free frozen parameters, and 3) fixed high-timestep denoising.

Next, even under these stringent constraints, we propose a simple diagnostic method, \textbf{AntiPure}, which achieves an effective protective outcome by incorporating two additional types of guidance: \textbf{Patch-wise Frequency Guidance (PFG)} and \textbf{Erroneous Timestep Guidance (ETG)}. 
Since the priors on clean images embedded in frozen parameters prioritize low-frequency structures, the purification model lacks fine control over high-frequency details. 
Thus, PFG modulates the high-frequency components in each patch of the model's predicted ``clean" image, leading to more localized perceptual discrepancy introduced by perturbation injection.
Moreover, while overall structures are anchored by high-timestep denoising, ETG helps circumvent the timestep limitation. By minimizing the output distance across timesteps, ETG reduces the model’s sensitivity to its timestep input and hinders its capacity to determine the appropriate actions at each step.
Together, these mechanisms empower AntiPure to achieve both minimal perceptual discrepancy and maximal output distortion.

In brief, our contributions can be summarized as follows:
\begin{itemize}
\item We first formalize the requirements for effective protective perturbations in the P-C workflow and propose anti-purification to target purification directly with a thorough analysis of the core challenges, overcoming the limitations of prior anti-customization perturbations.
\item We propose a simple diagnostic method named AntiPure. AntiPure leverages Patch-wise Frequency Guidance and Erroneous Timestep Guidance to counter the aforementioned anti-purification challenges. 
\item Finally, we establish a benchmark to assess the effectiveness of prior anti-customization methods and our AntiPure within the P-C workflow.
Experimental results show that AntiPure achieves both the lowest perceptual discrepancy and the highest output distortion as purification converges, outperforming existing methods.
\end{itemize}

\section{Related Works}
\label{sec:relatedwork}
\textbf{Customization with Stable Diffusion.}
With the rapid development of diffusion model \cite{sohl2015deep,ho2020denoising,song2020denoising,song2019generative,song2020score,dhariwal2021diffusion,ho2022classifier}, a series of Text-to-Image (T2I) models \cite{rombach2022high,ramesh2021zero,ramesh2022hierarchical,saharia2022photorealistic,nichol2021glide} show exceptional potential for customized generation, where open-source Stable Diffusion (SD) \cite{rombach2022high} emerges as a community favorite. By various fine-tuning and conditional control techniques, pretrained models can further meet the needs for specific concepts and finer control. 
Textual Inversion \cite{gal2022image} learns pseudo-words in the T2I models' embedding space. 
DreamBooth \cite{ruiz2023dreambooth} incorporates class-specific prior preservation loss during full fine-tuning to mitigate forgetting. 
Following parameter-efficient fine-tuning (PEFT) \cite{houlsby2019parameter}, Custom Diffusion \cite{kumari2023multi} modifies only the weights of cross-attention layers, while LoRA \cite{hu2021lora}, adapted from large language models (LLMs), views concepts as the offset of parameters using rank decomposition matrices. 
For extra condition control, T2I-Adapter \cite{mou2024t2i} and ControlNet \cite{zhang2023adding} integrate additional guidance from other conditioning inputs.

\noindent \textbf{Anti-Customization with Protective Perturbations.}
Advances in customization have raised significant concerns about deepfakes, privacy, and copyright. 
In this context, protective perturbation can serve as a potential solution to help prevent misuse and ensure authenticity. 
This can date back to the era of Generative Adversarial Networks (GANs) \cite{goodfellow2014generative}, when studies \cite{yeh2020disrupting,ruiz2020disrupting,wang2020deceiving,huang2022cmua} explored to distort the outputs of GAN-based image translation and editing \cite{zhu2017unpaired,choi2018stargan,wang2018high,he2019attgan} by white-box attacks \cite{goodfellow2014explaining,kurakin2016adversarial,kurakin2018adversarial,madry2017towards}.

For diffusion models, AdvDM \cite{liang2023adversarial} firstly employs Projected Gradient Descent (PGD) \cite{madry2017towards} to maximize the Latent Diffusion Model's training loss \cite{wang2018high} using Monte Carlo estimation. This method is further extended in Mist \cite{liang2023mist} with the addition of textural loss. Glaze \cite{shan2023glaze} safeguards artwork by attacking SD's encoder, while PhotoGuard \cite{salman2023raising} addresses unauthorized image inpainting by targeting both the encoder and UNet. 
Zhu et al. \cite{zhu2024watermark} utilize a GAN-based generator to create adversarial examples embedded with traceable watermarks. Anti-DreamBooth (Anti-DB) \cite{van2023anti} targets the fine-tuning process in DreamBooth by introducing a novel backpropagation surrogate to learn from both clean and partially adversarial examples. Building upon Anti-DB, SimAC \cite{wang2024simac} devises an adaptive greedy time interval selection. Zhao et al. \cite{zhao2024can} summarize the current challenges, providing a benchmark across fine-tuning methods and revealing the susceptibility of current techniques to purification. 
Concurrently, MetaCloak \cite{liu2024metacloak} meta-learns transformation-robust, transferable protections, while CAAT \cite{xu2024perturbing} perturbs cross-attention to obtain efficient, training-free perturbations. 
More recently, FastProtect \cite{ahn2025nearly} targets real-time deployment, and IDProtector \cite{song2025idprotector} trains a one-pass encoder to defend against generation.

\noindent \textbf{Adversarial Purification with Diffusion Models.} 
Purification removes adversarial noises by regenerating or refining input images, with diffusion models gaining attention for their iterative denoising abilities \cite{wang2022guided, nie2022diffusion, xiao2023densepure}. 
DiffPure \cite{nie2022diffusion} employs an unconditional diffusion model to diffuse the adversarial sample over a selected timestep and then denoises it by solving the reverse-time SDE. 
DensePure \cite{xiao2023densepure} enhances the certified robustness of the pretrained classifier. However, these classifier-focused methods may not be sensitive to the high-resolution and perceptual consistency required for customization. 
GrIDPure \cite{zhao2024can} employs shorter timesteps diffusion with multiple iterations and overlapping grids to improve purification for customization.

\section{Preliminaries}
Here, we briefly introduce the necessary preliminaries. For more details, please refer to Appendix \textcolor{iccvblue}{A}.

\noindent \textbf{Customization.} The essence of customization is to fine-tune a model, pretrained on large-scale data, on a smaller, concept-specific set to capture that unseen concept, and Stable Diffusion (SD) \cite{rombach2022high} is a popular choice.
Given input image $x_0$ and its text prompt $y$, the noise predictor UNet \cite{ronneberger2015u} $\epsilon_{\theta_c}$ and the text encoder $\tau_{\theta_c}$ with SD's parameters $\theta_c$ for customization are jointly optimized through:
\begin{equation}
\mathcal{L}_{ldm}(x_0; \theta_c) = \mathbb{E}_{ \epsilon \sim \mathcal{N}(0,\textbf{I}), t \sim \mathcal{U}(1,T)} 
\left\| \epsilon - \epsilon_{\theta_c}(z_t, t, \tau_{\theta_c}(y)) \right\|^2_2,
\label{eq:condLDM}
\end{equation}
where latent $z_t$ is sampled through closed-form diffusion under reparameterization given VAE encoded $z_0 = \mathcal{E}(x_0)$. 

\noindent \textbf{Anti-customization} utilizes adversarial attacks against generation, aiming to distort the concepts learned during fine-tuning by injecting protective perturbation $\delta^{adv}$. For the optimal solution, this presents a saddle point problem:
\begin{equation}
\delta^{adv} = \underset{\|\delta\|_\infty \leq \eta}{\arg\max} \, \underset{\theta_c} {\min} \, \mathbb{E}_{x} \mathcal{L}_{ldm}(x_0 + \delta; \theta_c),
\label{eq:adv}
\end{equation}
where $\delta^{adv} = x^{adv} - x_0$ is the adversarial perturbation restricted within the $l_\infty$-ball of radius $\eta$. In practice, we often simplify \cref{eq:adv} and maximize $\mathcal{L}_{ldm}$ to approximate the optimal $\delta^{adv}$ by white-box methods like I-FGSM \cite{kurakin2016adversarial,kurakin2018adversarial} or Projected Gradient Descent (PGD) \cite{madry2017towards}. 

\noindent \textbf{Purification} can remove adversarial noises while maintaining the global structures, thereby rendering protective perturbations ineffective. 
Pretrained unconditional diffusion models, such as DDPMs \cite{ho2020denoising}, can be inherently used for purification since the distributions of clean and adversarial samples converge over time during forward diffusion. 

The pioneering work of diffusion-based purification, DiffPure \cite{nie2022diffusion}, diffuses the input adversarial image at timestep $t^{p}$ and denoises it back to a purified image. In simplified discrete DDPM form, this can be written as:
\begin{equation}
\operatorname{Pure}(x^{adv})  = \operatorname{Reverse}(\sqrt{\overline{\alpha}_{t^{p}}} (x^{adv}) + \sqrt{1- \overline{\alpha}_{t^{p}}} \epsilon, t^{p}, 0; \theta_p),
\label{eq:pure}
\end{equation}
where $\overline{\alpha}_{t^p} := \prod _{t=1}^{t^p}(1-\beta_{t})$ with $\beta_{t}$ representing the diffusion variance. $\operatorname{Reserve(\cdot)}$ iteratively denoises the adversarial sample diffused at timestep $t^{p}$ from higher timestep $t^{p}$ to lower timestep $0$ via learning-free sampling with frozen purification parameters $\theta_p$.
In the field of anti-customization, GrIDPure \cite{zhao2024can} adapts DiffPure to meet SD's requirements, transforming the longer denoising timestep into multiple shorter iterations, and achieving enhanced purified results. 

\section{Analysis}
\label{sec:analysis}
\subsection{Anti-purification: Overall Formulation}\label{sec:formulas}
For ideal perturbations resistant to purification, we first formalize our objective as follows:
\begin{equation}
\delta^{adv} = \underset{\|\delta\|_\infty \leq \eta}{\arg\max} \, \underset{\theta_c} {\min} \, \mathbb{E}_{x} \mathcal{L}_{ldm}(\operatorname{Pure}(x_0 + \delta); \theta_{c}),
\label{eq:robust}
\end{equation}
where $\delta^{adv}$ is similarly optimized by maximizing $\mathcal{L}_{ldm}$, but with the purified input substituting the original one.
However, direct backpropagation is computationally inefficient here, as purification generates extremely deep computation graphs over multiple iterations. 
Besides, unlike conventional adversarial attacks, anti-customization requires substantial massive memory overhead associated with SD, to which the size of classifiers is not comparable.

Alternatively, we decompose \cref{eq:robust} into stages. Because the P-C workflow is serial, as long as one link in the procedure fails, effective fine-tuning cannot be realized. Interestingly, two opposing objectives can accomplish this:
\begin{equation}
\delta^{adv\prime}_{min} = \underset{\|\delta\|_\infty \leq \eta}{\arg\min} \, \| \operatorname{Pure}(x_0 + \delta) - (x_0 + \delta)\|_\infty, \ \text{or}
\label{eq:min}
\end{equation}
\begin{equation}
\delta^{adv\prime}_{max} = \underset{\|\delta\|_\infty \leq \eta}{\arg\max} \, \| \operatorname{Pure}(x_0 + \delta) - (x_0 + \delta)\|_\infty.
\label{eq:max}
\end{equation}

\begin{figure}[t]
\centering
\includegraphics[width=0.8\linewidth]{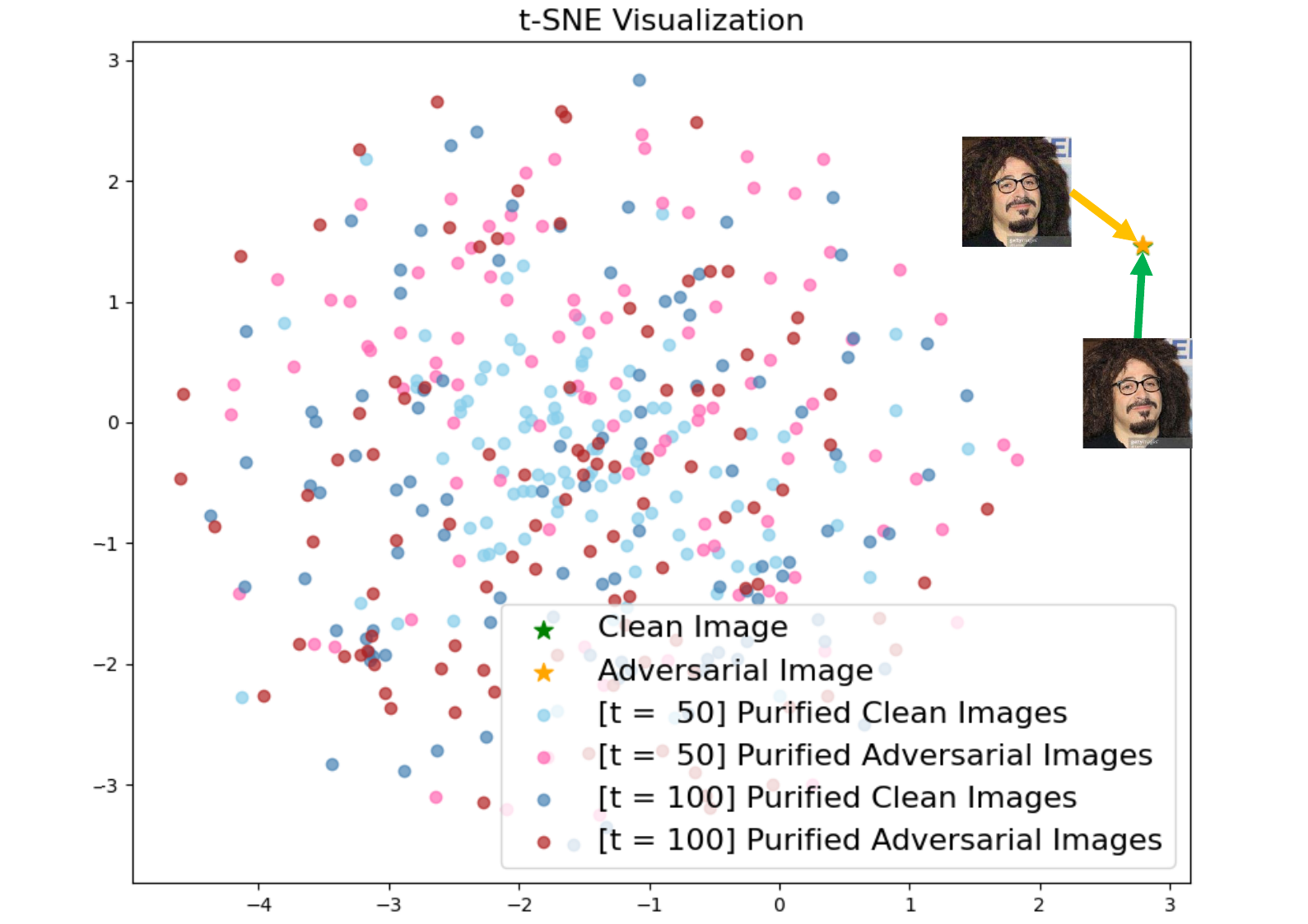}
\vspace{-5pt}
\caption{t-SNE \cite{van2008visualizing} visualization (perplexity = 10) of 4$\times$100 purified images obtained using DiffPure \cite{nie2022diffusion} with different timesteps for clean and adversarial images.}
\vspace{-10pt}
\label{fig:t-sne}
\end{figure}

By \cref{eq:min} combined with \cref{eq:adv}, we can approximate $\operatorname{Pure}(x) \approx x$, allowing \cref{eq:robust} to degenerate into \cref{eq:adv} even under purification. This follows the principle of adaptive attack and leads to a robust perturbation that can still target customization after purification.  
However, involving joint optimization of two objectives (\cref{eq:min} and \cref{eq:adv}) makes the optimal solution hard to find as well. 
Additionally, probabilistic diffusion models may result in generated images that are almost completely uncontrollable at the fine scale required for adversarial noise. 
As shown in \cref{fig:t-sne}, the difference between the clean and adversarial images (which nearly overlap) is far smaller than the range of purified outputs, and the distributions of the purified clean and adversarial images converge as $t^{p}$ increases. 
Therefore, we advocate for the alternative approach instead of treating purification as a special transformation and requiring the purified outputs to precisely reach the desired ``adversarial region." 

By \cref{eq:max}, we resort to direct attacks against purification, i.e., \textbf{anti-purification}. We note that anti-purification is not intended as a wholesale replacement for anti-customization, but rather complements it as a preparatory step within diverse emerging workflows. In such scenarios, even if subsequent customization operates normally, the target concepts learned are still distorted from those of the original images during purification. This shifts the focus of our attack from the customization's LDMs to the DDPMs used in (DDPM-based) purification:
\begin{equation}
\mathcal{L}_{ddpm}(x_0; \theta_p) = \mathbb{E}_{ \epsilon \sim \mathcal{N}(0,\textbf{I}), t \sim \mathcal{U}(1,T)} 
\left\| \epsilon - \epsilon_{\theta_p}(x_t, t) \right\|^2_2.
\label{eq:ddpm}
\end{equation}
Unfortunately, direct attacks via maximizing this DDPM training loss cannot achieve the same level of semantic structural distortion seen in anti-customization methods due to the inherent characteristics of purification itself. 
This makes anti-purification a more challenging task, with the difficulties encapsulated in the following three core reasons.

\subsection{Anti-purification: Why harder?}\label{sec:why}
Here, we report the conclusions of our small-scale experiments. For detailed settings, please refer to Appendix \textcolor{iccvblue}{B.2}.

\subsubsection{Reason 1:  Lack of Vulnerable Components}\label{sec:r1}
Attacks targeting LDMs/SD are easier due to their more vulnerable encoders. In contrast, the only component in DDPMs, the UNet \cite{ronneberger2015u}, is extremely robust. 

\begin{figure}[t]
\centering
\includegraphics[width=0.8\linewidth]{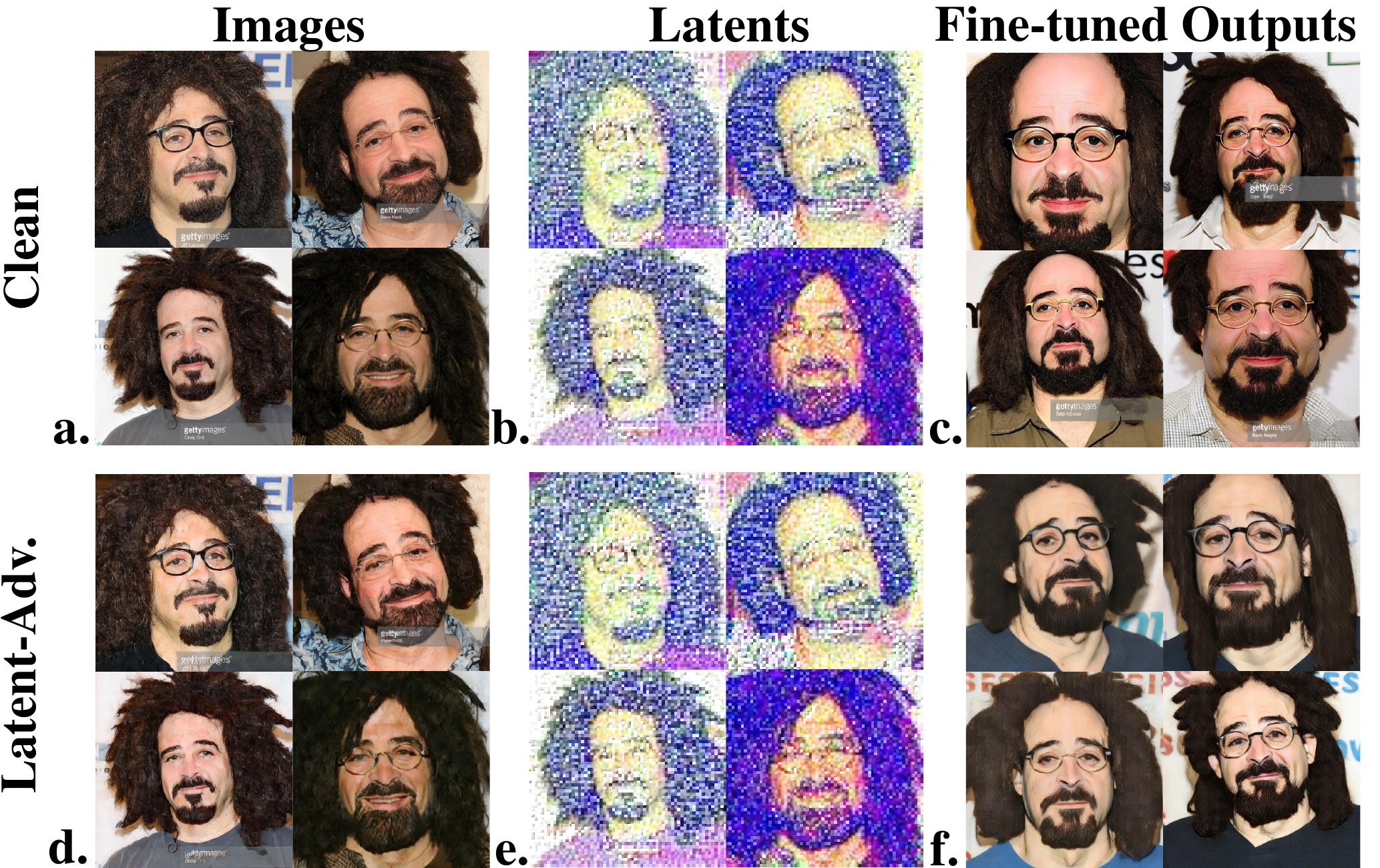}
\caption{Attacks against DreamBooth \cite{ruiz2023dreambooth} on UNet are much harder. Unlike vanilla pixel-space attacks ($a. \rightarrow d.$), latent-space attacks ($b.\rightarrow e.$) cannot target the vulnerable VAE encoder. Here, $d.$ (decoded from $e.$) is shown for visualization purposes only; in our experiments, we directly replace $b.$ with $e.$ during fine-tuning.} 
\label{fig:zatt}
\end{figure}
Firstly, we modify Anti-DB's ASPL method \cite{van2023anti} to conduct PGD attacks directly on \cref{eq:condLDM} but in the latent space, thereby obtaining ($e.$) adversarial latents in \cref{fig:zatt} (rather than adversarial images for further encoding). As shown in \cref{fig:zatt}, ($d.$) the decoded images from these latents exhibit unexpected stylistic transformations, and ($f.$) images generated by SD fine-tuned on these latents show minimal differences from ($d.$). This indicates that if PGD cannot leverage gradients from the vulnerable VAE encoder, the effectiveness of attacks is significantly reduced. 

\begin{figure}[t]
\centering
\includegraphics[width=0.9\linewidth]{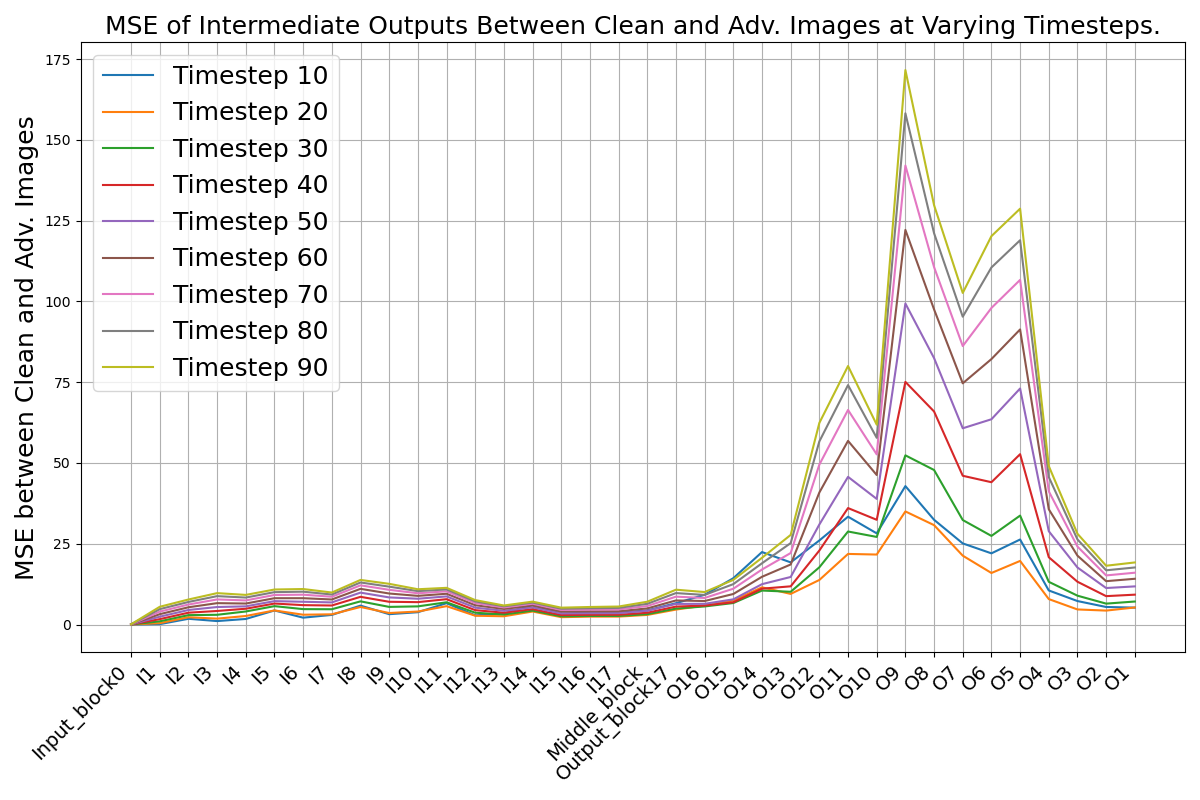}
\vspace{-5pt}
\caption{Mean Squared Errors of intermediate outputs between clean and adversarial samples across different UNet blocks at varying timesteps. See Appendix \textcolor{iccvblue}{B.2} for details.} 
\vspace{-10pt}
\label{fig:output}
\end{figure}

Also, we directly attack $\mathcal{L}_{ddpm}$. \cref{fig:output} illustrates the differences between clean and adversarial images in the outputs of various UNet blocks at different timesteps. Notice that there is minimal difference in the downsampling and intermediate blocks, while substantial deviations emerge in the middle part of the upsampling blocks. However, the outputs still converge at last, which may be attributed to the unique residual connections: the effect of adversarial noise naturally accumulates as the network's spatial structure iterates, while the shallower residual connections weaken this impact, forcing the perturbation to subside to lower levels, leading to a peak effect.

\subsubsection{Reason 2: Training-free Frozen Parameters}\label{sec:r2}
\begin{figure}[t]
\centering
\includegraphics[width=1.0\linewidth]{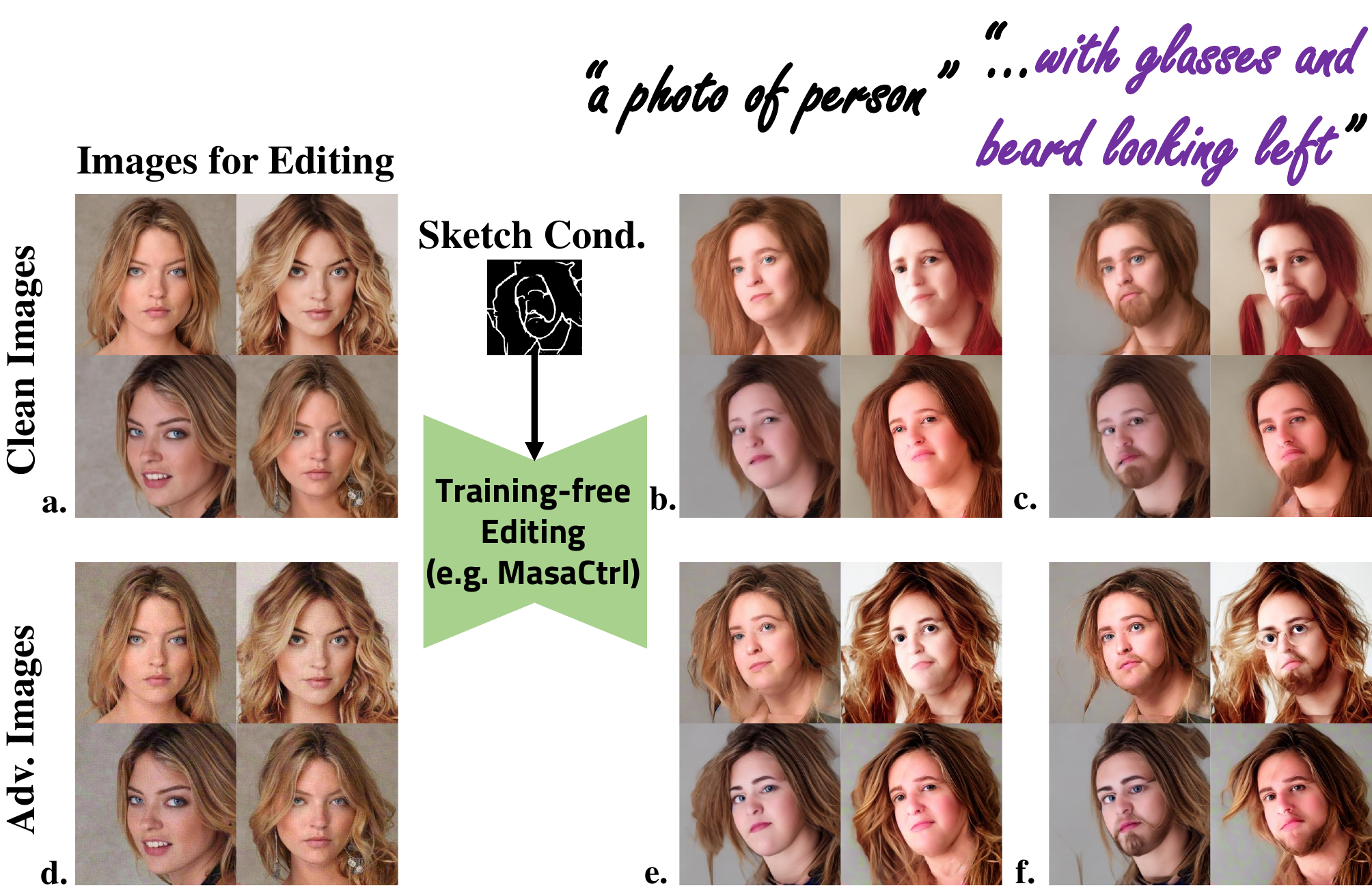}
\caption{Effectiveness of MasaCtrl \cite{cao2023masactrl} on adversarial images. The loss attack makes little difference except that the lower right image of $e.$ and $f.$ has slight noise in the background.} 
\vspace{-10pt}
\label{fig:masa}
\end{figure}

It is both evident and important to note that, unlike anti-customization which targets fine-tuning by data poisoning, anti-purification targets a training-free editing task. In anti-purification, the pretrained purification model requires no additional adjustments, so adversarial samples cannot affect its benign priors embedded in frozen $\theta_p$. Similar issues also arise in attacks on other training-free editing tasks. As shown in \cref{fig:masa}, the perturbations can reduce image quality in some cases for MasaCtrl \cite{cao2023masactrl}, which also relies on vulnerable encoders. However, it does not achieve the same degradation level as perturbed fine-tuning.

\subsubsection{Reason 3: Fixed High Timestep Denoising}\label{sec:r3}
During the denoising process, the low-frequency structural information of an image is largely determined at higher timesteps, while denoising at lower timesteps focuses on high-frequency, textural changes. 
Here, we notice that \textbf{the purification process can be viewed as a generation process where high-timestep denoising is fixed}. In cases where vulnerable components are absent and parameters are frozen, conducting a $\mathcal{L}_{ddpm}$-based attack for timesteps beyond \(t^{p}\) is not directly meaningful, and attempting to achieve semantic structural changes by adjusting the input at low timesteps is also unfeasible. Essentially, attacks on the purification itself are attacks restricted to high-frequency components at low timesteps.

\section{Method: AntiPure}
\begin{figure}[t]
\centering
\includegraphics[width=1.0\linewidth]{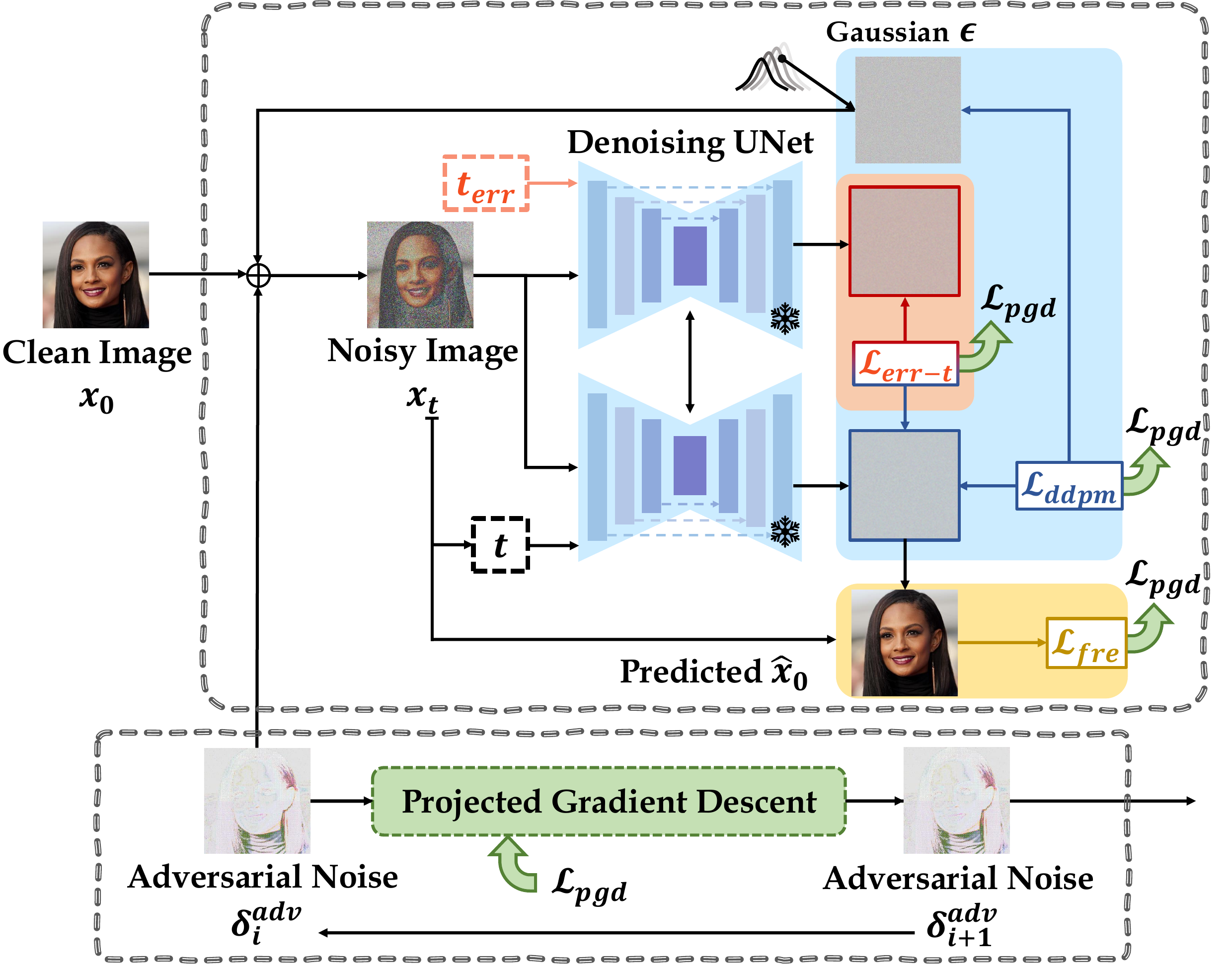}
\caption{Overview diagram of AntiPure. AntiPure mitigates the negative impact of benign priors (\cref{sec:r2}) and limited timesteps (\cref{sec:r3}) on the attack by introducing Patch-wise Frequency Guidance $\mathcal{L}_{fre}$ and Erroneous Timestep Guidance $\mathcal{L}_{t-err}$.} 
\vspace{-10pt}
\label{fig:antipure}
\end{figure}

Building on the formulation in \cref{sec:formulas}, we aim to generate the most adversarial input specifically targeting the purification model itself via \cref{eq:max}. 
As highlighted in \cref{sec:why}, achieving semantic structural distortion like anti-customization is unfeasible for anti-purification. 
However, we can still raise the costs of customization and achieve protective perturbations that make the outputs from the P-C workflow more distinguishable.
The overall attack process of our AntiPure is shown in \cref{fig:antipure}.

\label{sec:method}

\subsection{Patch-wise Frequency Guidance}
\begin{figure}[t]
\centering
\includegraphics[width=0.9\linewidth]{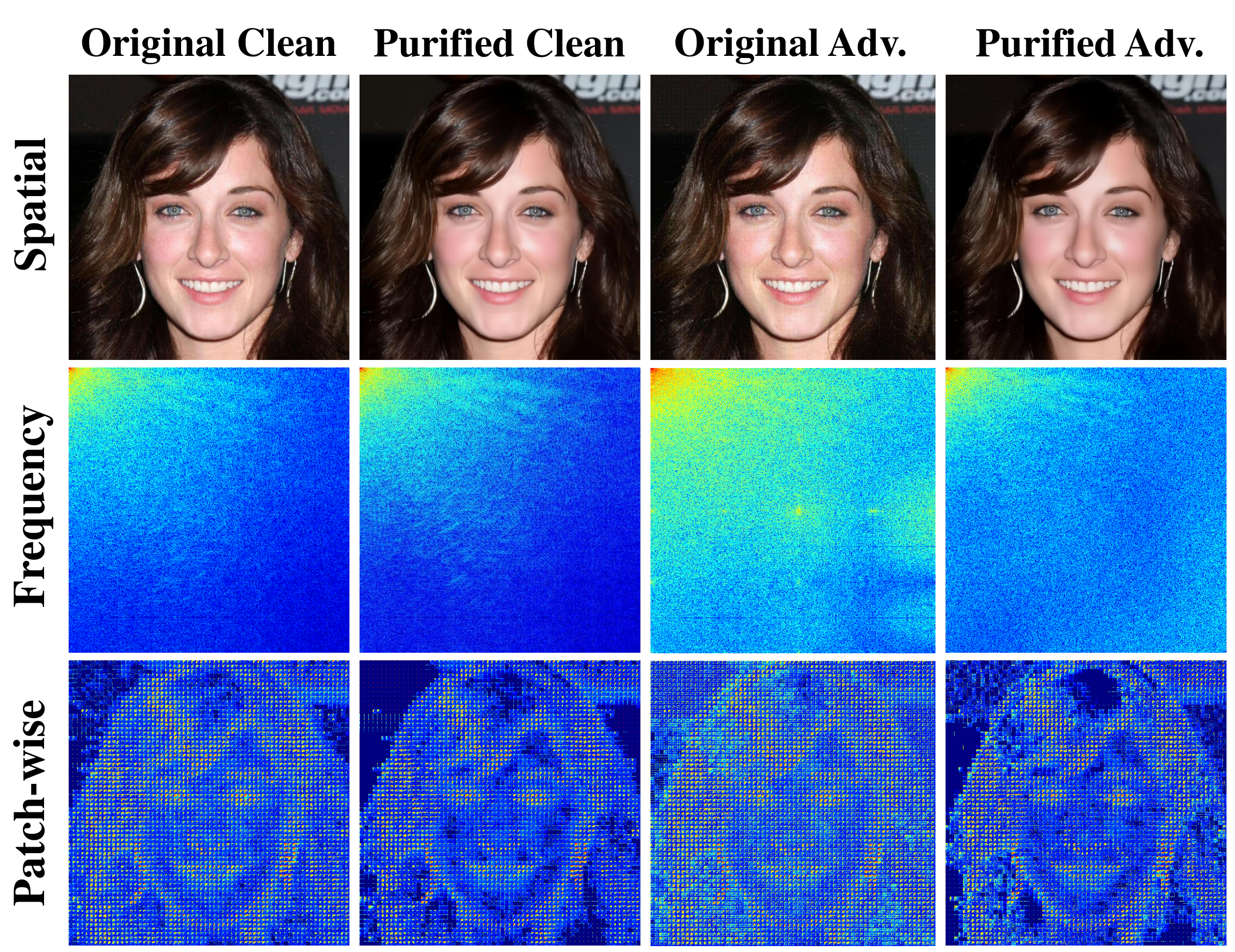}
\caption{Differences in the spatial and frequency domains before and after DDPM-purification. Pseudocolor transformation is applied to the DCT spectrogram for better visualization.} 
\vspace{-10pt}
\label{fig:dct}
\end{figure}

The priors on clean images embedded in the frozen network parameters allow the purification model to revert to a high-quality image distribution aligned with human intuition during the reverse denoising process.
The overall outputs can often remain structurally high-quality even if a particular timestep is effectively attacked.
However, unlike low-frequency semantic structures, consistency in high-frequency components is harder to guarantee, rendering them less controllable during purification.
As shown in \cref{fig:dct}, although simply using \(\mathcal{L}_{ddpm}\)-attacks does not produce noticeable effects in the spatial domain, they introduce observable discrepancies in the frequency domain.

This phenomenon motivates shifting our focus to the frequency domain.
A feasible solution is to modulate the high-frequency components where purification typically exerts weaker control, thereby causing the purified outputs to deviate from clean priors mainly in high-frequency bands.
Also, due to the characteristics of human perception, modulating high-frequency components affects image semantic information less.
For finer spatial modulation, we operate on smaller-resolution patches to localize perceptual discrepancy, forming our Patch-wise Frequency Guidance (PFG). 

Specifically, given a clean image $x_0\in \mathbb{R}^{C\times H \times W}$, a Gaussian noise $\epsilon$, and an adversarial noise $\delta^{adv}_i$ at the $i$-th step of PGD, we diffuse the noisy adversarial sample $x_t$ at timestep $t$ as follows:
\begin{equation}
x_t  = \sqrt{\overline{\alpha}_t} (x_0 + \delta^{adv}_i) + \sqrt{1- \overline{\alpha}_t} \epsilon,
\label{eq:xt}
\end{equation}
where $\overline{\alpha}_t := \prod _{t'=1}^{t}(1-\beta_{t'})$ with $\beta_{t}$ representing the diffusion variance. Using the UNet's output $\epsilon_\theta(x_t, t)$, we approximate the predicted denoised image $\widehat{x}_0$ by:
\begin{equation}
\widehat{x}_0 = (x_t - \sqrt{1- \overline{\alpha}_t} \epsilon_\theta(x_t, t)) / \sqrt{\overline{\alpha}_t}.
\label{eq:x0}
\end{equation}
Next, Patch-wise Frequency Guidance operates on $\widehat{x}_0$ to track the UNet's gradients while emphasizing high-frequency components, which can be formalized as:
\begin{equation}
\mathcal{L}_{fre}(x_0;\delta^{adv})  = \sigma ( \mathbb{E}_{P} \frac{4}{s^2} \sum \limits_{m,n=\frac{s}{2}}^{s-1} \operatorname{PatchDCT} (\widehat{x}_0, s)_{m,n} ),
\label{eq:lfre}
\end{equation}
where $\operatorname{PatchDCT}(\cdot)$ (illustrated in $3^{rd}$ row of \cref{fig:dct}) unfolds $\widehat{x}_0$ into $P$ patches of size $s \times s$, and applies the Discrete Cosine Transform (DCT) to each patch. After DCT, the low-frequency component occupies the top-left corner of the spectrogram, while the high-frequency component lies in the bottom-right corner. Consequently, the bottom-right quarters of each patch are filtered, averaged, and then sigmoid-normalized by $\sigma(\cdot)$ to yield $\mathcal{L}_{fre}$. 

In brief, $\mathcal{L}_{fre}$ aims to enhance the high-frequency components of the purification model’s prediction after denoising across different timesteps, indirectly reinforcing the adversarial perturbation’s high-frequency elements and creating a uniform grid pattern.
Since the attack targets higher frequencies, localized structural information is minimally altered, ensuring greater perceptual consistency for humans.

\subsection{Erroneous Timestep Guidance}
As mentioned in \cref{sec:r3}, purification can essentially be viewed as a generative process where the high-timestep denoising has been fixed, meaning that the structure of images cannot be obviously altered. However, we can encourage the noise predictor’s outputs at distinct timesteps to remain as close as possible. By injecting adversarial noise, we identify inputs for which the UNet struggles to select the appropriate actions across timesteps, which we realize via Erroneous Timestep Guidance (ETG).

Specifically, we select an erroneous timestep $t_{err}$ as the input to the UNet to obtain a noise prediction of \( x_t \) at a higher timestep, and we minimize the discrepancy between the noise predicted at the erroneous time step $\epsilon_\theta(x_t, t_{err})$ and the correct prediction $\epsilon_\theta(x_t, t)$:
\begin{equation}
\mathcal{L}_{err-t}(x_0;\delta^{adv})  = -\left\| \epsilon_\theta(x_t, t_{err}) - \epsilon_\theta(x_t, t)\right\|^2_2.
\label{eq:lt}
\end{equation}

\subsection{Overall Attack}
As shown in \cref{fig:antipure}, our attack incorporates $\mathcal{L}_{fre}$and $\mathcal{L}_{err-t}$ guidance with vanilla $\mathcal{L}_{ddpm}$. To balance the range of values across different training objectives, the total loss used for gradient ascent in PGD can be formalized as:
\begin{equation}
\begin{split}
\mathcal{L}_{pgd}(x_0;\delta^{adv})  = \mathbb{E}_{\epsilon, t}
\left( \mathcal{L}_{ddpm} + \lambda_1 e^{\overline{\alpha}_t - 1} \mathcal{L}_{fre} + \lambda_2  e^{\mathcal{L}_{err-t}} \right),
\end{split}
\label{eq:lall}
\end{equation}
where the attack timestep $t \sim \mathcal{U}(1,t^{p})$ is constrained within the purification step range $t^{p}$ to avoid ineffective attacks. Both hyperparameters \( \lambda_1 \) and \( \lambda_2 \) are set to 0.5, with \( t_{err} \) fixed for convenience. The coefficient $e^{\overline{\alpha}_t - 1}$ scales $\mathcal{L}_{fre}$ to increase its impact as $t$ decreases. Given the minimal MSE values, we apply an exponential function to $\mathcal{L}_{err-t}$ for slightly more aggressive optimization. Finally, PGD maximizes $\mathcal{L}_{pgd}$ to find the optimal $\delta^{adv}$.

\section{Experiments}
\label{sec:experiment}

\subsection{Experimental Setup}
\textbf{Dataset.} We utilize two facial datasets, CelebA-HQ \cite{karras2017progressive} and VGGFace2 \cite{cao2018vggface2} for evaluation. Following Anti-DB \cite{van2023anti}, we select two subsets on these two datasets, each containing 50 IDs with 12 images at 512$\times$512 resolution per ID, to ensure training for both Anti-DB and SimAC \cite{wang2024simac}.

\noindent \textbf{Baseline.} We apply four advanced protective perturbation methods, including AdvDM \cite{liang2023adversarial}, Mist \cite{liang2023mist}, Anti-DB, and SimAC, where Anti-DB and SimAC require additional instances for perturbation generation. Thus, for each ID, we generate perturbation for 4 images and use another 4 images as instances for Anti-DB/SimAC.

\noindent \textbf{Quantitative Metrics.} For customization's output, we first employ \textbf{FID} \cite{heusel2017gans}, a general metric in generative tasks that quantifies the distance between the feature distributions of synthetic and real images using Inception v3 \cite{szegedy2016rethinking}. 
Additionally, we include metrics from Anti-DB \cite{van2023anti}.  Since our primary goal is to generate non-realistic faces, we employ \textbf{Identity Score Matching (ISM)} to evaluate the cosine similarity between the synthetic image's facial features and the real features of the corresponding identity using the popular ArcFace recognizer \cite{deng2019arcface}. We also report Face Detection Failure Rate (FDFR) using the RetinaFace detector \cite{deng2020retinaface}, measuring the ratio of undetectable faces excluded in ISM evaluation. This metric merely serves to provide ISM with a more comprehensive reference, as producing completely distorted faces after purification is, as previously discussed, unfeasible. 
Finally, we use an extra no-reference image quality assessment metric, \textbf{BRISQUE} \cite{mittal2012no}, a classical and popular method for measuring generated image quality.
For perturbation itself, a reconstruction metric, \textbf{LPIPS} \cite{zhang2018unreasonable} is provided to evaluate the difference in human perception.
As we focus on disrupting the outputs of SD after purification, higher FID, lower ISM, and higher BRISQUE mean an increase in the adversarial effect. Besides, lower LPIPS means more imperceptible perturbation.

\noindent \textbf{Configurations.} For all perturbation methods, we use the same PGD training/purification/customization setup. In \cref{sec:db} and \cref{sec:lora}, we use GrIDPure \cite{zhao2024can} for purification, applying 2 rounds of 20 iterations with $t^{p} = 10$ to approximate convergence of the purification effect (validated in \cref{sec:pure}). 
In \cref{sec:percpt}, we evaluate perceptual consistency with pretrained AlexNet\cite{krizhevsky2012imagenet}/VGG\cite{simonyan2014very}. 
To verify the robustness of our method against common image processing techniques, we save the images with protective perturbations in their original input format, that is, the results on CelebA-HQ (.jpg) are all JPEG-compressed.
We will elaborate on our experimental configurations in Appendix \textcolor{iccvblue}{C}.

Furthermore, for ablation studies on different perturbation configurations, black-box performance, and more visualization results, please refer to Appendix \textcolor{iccvblue}{D}.



\subsection{Comparison with Baseline Methods}\label{sec:db}
\begin{table}[t]
\renewcommand{\arraystretch}{1.5}
\centering
\resizebox{\columnwidth}{!}{
\begin{tabular}{ccccc}
\hline
Dataset & Perturbation & FID↑ & ISM↓ (FDFR) & BRISQUE↑ \\ \hline

\multirow{5}{*}{CelebA-HQ}& AdvDM \cite{liang2023adversarial}& 77.51 & 0.6561 (0.10) & 31.33\\ 
& Mist \cite{liang2023mist}&  70.23 &  0.6688 (0.07) &  37.00\\ 
& Anti-DB \cite{van2023anti}& 78.84 & 0.6422 (0.10) & 31.76\\ 
& SimAC \cite{wang2024simac}& 67.37 & 0.6734 (0.09) & 33.73\\ \cline{2-5} 
& \textbf{AntiPure (Ours)}& \textbf{81.15} & \textbf{0.6112} (0.10) & \textbf{43.60}\\ \hline

\multirow{5}{*}{VGGFace2}& AdvDM \cite{liang2023adversarial}& 83.90 & 0.5923 (0.09) & 37.42\\ 
& Mist \cite{liang2023mist}&  78.34 &  0.5940 (0.07)&  43.60\\ 
& Anti-DB \cite{van2023anti}& 90.29 & 0.5938 (0.06)& 38.35\\ 
& SimAC \cite{wang2024simac}& 75.21& 0.6053 (0.09)& 40.27\\ \cline{2-5} 
& \textbf{AntiPure (Ours)}& \textbf{90.77}& \textbf{0.5475} (0.05) & \textbf{46.01}\\ \hline
\end{tabular}
}
\caption{Comparison of DreamBooth's \cite{ruiz2023dreambooth} output quality for different perturbation methods following the P-C workflow.}
\label{tab:db}
\vspace{-10pt}
\end{table}

\noindent \textbf{Quantitative Results.} We first fine-tune SD on the purified images with different perturbations by DreamBooth, setting both the instance and inference prompt to ``a photo of \textit{sks} person." For fairness, the same set of 200 prior class images is shared between all perturbation methods. These quantitative results are shown in \cref{tab:db}.

In \cref{tab:db}, AntiPure achieves the best performance across all metrics on both datasets. After sufficient purification, the effectiveness of the previous methods is significantly reduced, but the effect of AntiPure can be better preserved. More importantly, while these methods depend on various customization configurations, AntiPure focuses solely on the relatively simpler purification process. 
On the other hand, due to the perturbed image degradation caused by JPEG compression, the results on CelebA-HQ are noticeably worse than those on VGGFace2, even when accounting for potential domain gaps.

Interestingly, SimAC shows an even greater drop in performance after purification. Compared to Anti-DB, the improved SimAC generates smoother edges around the perturbation patterns, rendering them more vulnerable. A similar distinction is also noted between AdvDM and Mist.

\begin{figure}[t]
\centering
\includegraphics[width=0.95\linewidth]{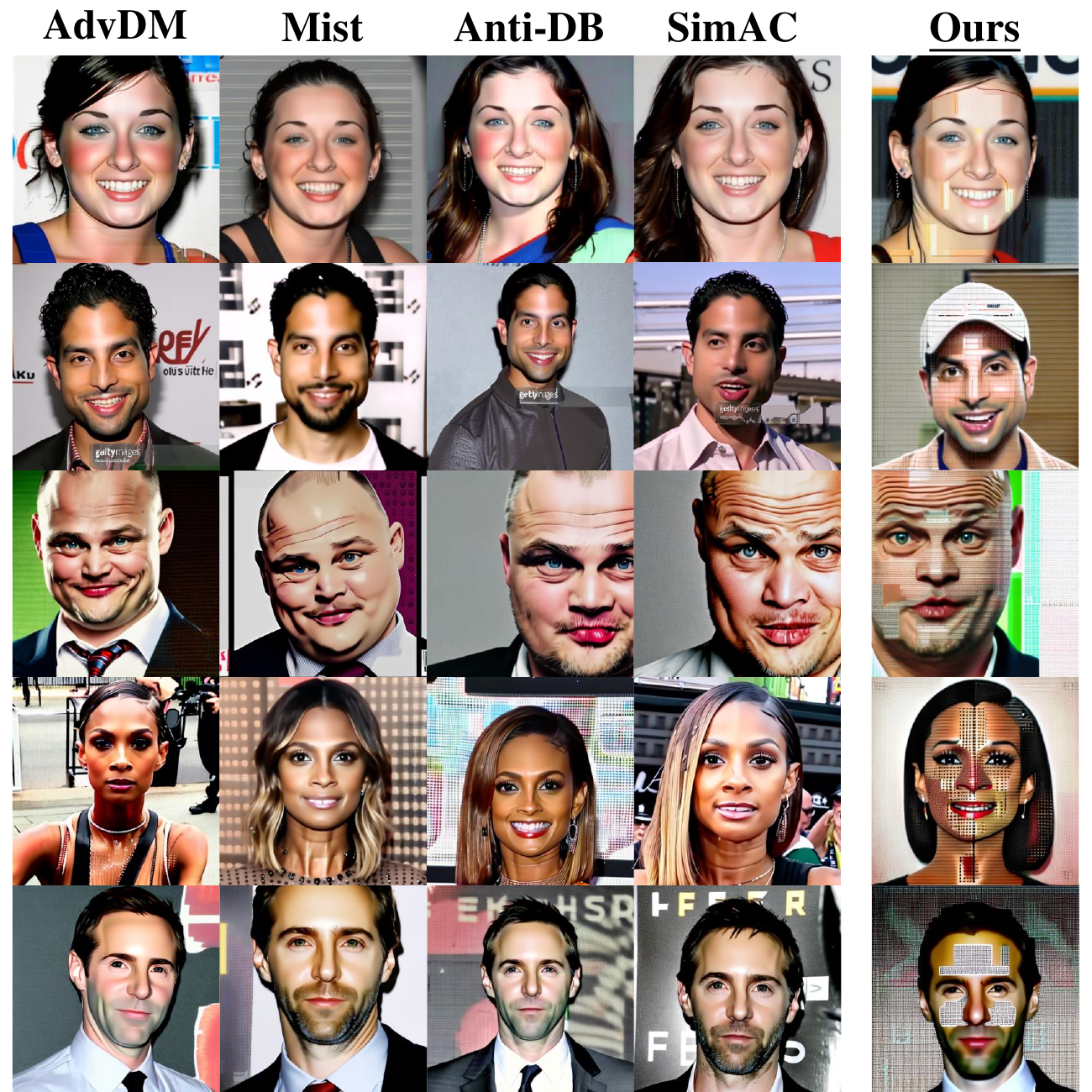}
\caption{Visualization of DreamBooth's outputs after the P-C workflow.} 
\vspace{-15pt}
\label{fig:more}
\end{figure}

\noindent \textbf{Qualitative Results.}
We show more results in \cref{fig:more}. In our selected examples, the degree of output degradation intensifies from top to bottom. As previously discussed, after purification, perturbation methods cannot realistically achieve distortion at the level of overall semantic structure. AntiPure, however, does not aim to adjust the perturbations to disrupt the customization process to the maximum extent. Instead, it simply aims for the purification outputs to degrade, leveraging the characteristics of the fine-tuning process itself to amplify these artifacts. This approach may not achieve distortion of a person’s ID information, but it is sufficient for humans to discern authenticity. For more visualization results, please refer to Appendix \textcolor{iccvblue}{D}.

\subsection{Different Customization Method}\label{sec:lora}
\begin{table}[t]
\renewcommand{\arraystretch}{1.5}
\centering
\resizebox{\columnwidth}{!}{
\begin{tabular}{ccccc}
\hline
Dataset & Workflow & FID↑ & ISM↓ (FDFR) & BRISQUE↑ \\ \hline

\multirow{4}{*}{CelebA-HQ} 
& AdvDM \cite{liang2023adversarial}     & 95.38 & 0.6302 (0.09)& 38.20\\ 
& Mist \cite{liang2023mist}     &  85.09 &  0.6461 (0.07)&  \textbf{40.91}\\ 
& Anti-DB \cite{van2023anti}   & 104.18 & 0.6215(0.12) & 38.18\\ 
& SimAC \cite{wang2024simac}     & 75.46 & 0.6487 (0.06) & 38.77\\ \cline{2-5} 
& \textbf{AntiPure (Ours)} & \textbf{109.63}& \textbf{0.5839} (0.07)& 40.01\\ \hline

\multirow{5}{*}{VGGFace2}& AdvDM \cite{liang2023adversarial}     & 105.43 & 0.5799 (0.07)& 58.02\\ 
& Mist \cite{liang2023mist}     &  90.66&  0.6046 (0.07)&  62.22\\ 
& Anti-DB \cite{van2023anti}   & 117.89& 0.5723 (0.06)& 58.56\\ 
& SimAC \cite{wang2024simac}     & 94.89& 0.6018 (0.07)& 59.99\\ \cline{2-5} 
& \textbf{AntiPure (Ours)} & \textbf{127.67}& \textbf{0.5428} (0.04)& \textbf{69.97}\\ \hline
\end{tabular}
}
\caption{Comparison of LoRA's \cite{hu2021lora} output image quality for different perturbation methods following the P-C workflow.}
\label{tab:lora}
\vspace{-10pt}
\end{table}
To evaluate AntiPure's generalizability, we also apply another popular fine-tuning method, LoRA. Unlike DreamBooth, LoRA significantly reduces fine-tuning costs through low-rank decomposition. It can meanwhile incorporate the core class-specific prior preservation loss of DreamBooth. As shown in \cref{tab:lora}, our method still outperforms other perturbation methods, with a noticeable gap in the ISM metric compared to others, indicating that our method effectively reduces facial feature similarity.

\subsection{Different Purification Configuration}\label{sec:pure}
Intuitively, as the purification iterations increase, protective perturbations are removed progressively, resulting in better fine-tuning output quality. To verify this, we select Anti-DB, the most robust method (apart from AntiPure) from previous experiments, and apply multiple rounds of purification with 10 iterations, followed by DreamBooth fine-tuning on CelebA-HQ. As shown in \ref{tab:pure}, the effectiveness of Anti-DB is indeed gradually diminished as the number of iterations increases, as expected, with ISM stabilizing around 20–40 iterations, indicating near convergence.

Unexpectedly, AntiPure, which focuses on countering the purification process, becomes increasingly robust with more iterations. Although its effectiveness is initially lower than that of Anti-DB with fewer iterations, AntiPure surpasses Anti-DB after approximately 30–40 iterations. This also demonstrates that the purification iteration settings in our previous experiments provide an accurate assessment of perturbation persistence at convergence.
\begin{table}[t]
\renewcommand{\arraystretch}{1.5}
\centering
\resizebox{\columnwidth}{!}{
\begin{tabular}{ccccc}
\hline
Perturbation & Workflow & FID↑  & ISM↓ (FDFR)& BRISQUE↑ \\ \hline

{None (Original)} & C (Iter=0) & 37.43 & 0.6935 (0.11) & 15.86\\ \hline

\multirow{4}{*}{Anti-DB \cite{van2023anti}} & P(Iter=10)-C  & 124.62 & 0.6020 (0.10) & 32.74\\ 
& P(Iter=20)-C   & 84.83 & 0.6352 (0.09)& 27.47\\ 
& P(Iter=30)-C   & 81.22 & 0.6473 (0.08)& 29.33\\  
& P(Iter=40)-C   & 77.30 & 0.6391 (0.09)& 30.34\\ \hline

\multirow{4}{*}{\textbf{AntiPure (Ours)}}& P(Iter=10)-C  & 54.45 & 0.6362 (0.07) & 40.27\\ 
& P(Iter=20)-C   & 59.97 & 0.6271 (0.08)& 44.63\\ 
& P(Iter=30)-C   & 68.84 & 0.6075 (0.08)& 47.68\\  
& P(Iter=40)-C   & 78.21 & 0.5994 (0.09)& 47.54\\ \hline
\end{tabular}
}
\caption{Comparison of DreamBooth's \cite{ruiz2023dreambooth} output image quality for different purification iterations following the P-C workflow on CelebA-HQ.}
\label{tab:pure}
\vspace{-5pt}
\end{table}

\subsection{Perceptual Consistency}\label{sec:percpt}

In addition to preventing infringement and forgery, protective perturbations also need to avoid affecting human perception, which is guaranteed by the hard constraint \(\eta\). However, even within the same \( l_\infty \)-ball, different perturbations can cause different perceptual shifts from clean images. To evaluate this, we select two commonly used backbones for LPIPS to assess the quality of different adversarial images under the same \(\eta\) constraint. As shown in \cref{tab:per}, AntiPure achieves the smallest perceptual difference while providing the best attacks against the P-C workflow. This can be attributed to Patch-wise Frequency Guidance, which effectively avoids modifications to low-frequency information.
\begin{table}[t]
\renewcommand{\arraystretch}{1.5}
\centering
\scriptsize
\setlength{\tabcolsep}{3pt}
{
\begin{tabular}{ccccc} \hline
\multirow{2}{*}{Perturbation} & \multicolumn{2}{c}{CelebA-HQ} & \multicolumn{2}{c}{VGGFace2} \\
\cmidrule(lr){2-3} \cmidrule(lr){4-5} & Alex-LPIPS↓& VGG-LPIPS↓ & Alex-LPIPS↓ & VGG-LPIPS↓ \\ \hline
AdvDM \cite{liang2023adversarial} & 0.2024 & 0.3061 & 0.2343 & 0.3920 \\
Mist \cite{liang2023mist} & 0.1470 & \textbf{0.2759} & 0.2208 & 0.5222 \\
Anti-DB \cite{van2023anti} & 0.2019 & 0.3319 & 0.2726 & 0.4054 \\
SimAC \cite{wang2024simac} & 0.1754 & 0.3046 & 0.2146 & 0.4120 \\ \hline
\textbf{AntiPure (Ours)} & \textbf{0.1392} & 0.2843 & \textbf{0.1758} & \textbf{0.3884} \\ \hline
\end{tabular}
}
\caption{Comparison of Learned Perceptual Image Patch Similarity (LPIPS) \cite{zhang2018unreasonable} between adversarial images obtained by different perturbation methods and the original images.}
\label{tab:per}
\vspace{-10pt}
\end{table}
\section{Conclusion}
In this paper, we formalize anti-purification in the P–C workflow and present AntiPure, which incorporates two types of additional guidance. Across representative P-C settings, AntiPure remains effective despite the strong denoising capability of diffusion-based purification, advancing perturbation robustness for stronger prevention against deepfakes and infringement.
However, as a simple diagnostic method, our current design assumes white-box access. Thus, we position AntiPure as a complement to anti-customization methods in diverse workflows, and leave improved black-box transfer as a direction for future work.

\section*{Acknowledgements}
This work is supported by Beijing Natural Science Foundation
(Grant No.~L252145); National Natural Science Foundation of China
(Grant Nos.~62425606, 32341009, U21B2045); the Strategic Priority
Research Program of Chinese Academy of Sciences
(Grant No.~XDA0480302); and the Young Scientists Fund of the State Key Laboratory of Multimodal Artificial Intelligence Systems (Grant No.~ES2P100116). 
We also acknowledge Zhida Zhang for his great
contributions to this work. His omission from the author list was an
inadvertent oversight during submission.

{
    \small
    \bibliographystyle{ieeenat_fullname}
    \bibliography{main}
}

\clearpage
\setcounter{page}{1}
\maketitlesupplementary

\renewcommand{\thesection}{\Alph{section}}
\renewcommand{\thesubsection}{\thesection.\arabic{subsection}}
\setcounter{section}{0}
\setcounter{table}{5}
\setcounter{figure}{9}
\setcounter{equation}{13}

\section{Details on Preliminaries} \label{sec:backsupp}
Here, we briefly review and supplement the preliminary knowledge from Sec. \textcolor{iccvblue}{3} to help our readers better understand the various tasks involved in this paper. \\

\noindent \textbf{Fine-tuning Customization (Personalization).}
Fine-tuning methods aim to inject specific concepts into the pretrained SD for customization. Among them, DreamBooth (DB) \cite{ruiz2023dreambooth}  is widely studied for anti-customization. This approach not only minimizes $\mathcal{L}_{ldm}$ in few-shot scenarios but also incorporates a prior-preservation term to retain beneficial prior knowledge, thus mitigating forgetting. Its training objective can be formalized as:
\begin{align}
\mathcal{L}_{db}(x_0; \theta_c) &= \mathcal{L}_{ldm}(x_0; \theta_c) \notag \\ 
& + \lambda \,\underbrace{\mathbb{E}_{\epsilon', t'} \|\epsilon' - \epsilon_{\theta_c}(z^{pr}_{t'}, t', \tau_{\theta_c}(y^{pr}))\|_2^2}_{\text{Class-Specific Prior Preservation Loss}},
\label{eq:dreambooth}
\end{align} 
where the class prior image $x^{pr}$ is generated by the pretrained model with class prompt $y^{pr}$, and $z^{pr}_0 = \mathcal{E}(x^{pr})$ diffuses at timestep $t'$ to form $z^{pr}_{t'}$  . In $\mathcal{L}_{db}$, the loss term $\mathcal{L}_{ldm}$ employs instance prompts $y$ of the form ``a photo of [V][class noun]," where [V] acts as an identifier describing the target concept. 

LoRA \cite{hu2021lora} is proposed to accelerate the optimization of large-scale pretrained models. It freezes the pretrained weights and injects trainable rank decomposition matrices into each layer, greatly reducing the number of trainable parameters for downstream tasks:
\begin{equation}
 W' = W_0 + A \cdot B,
 \label{eq:lora}
\end{equation}
where $W'$ and $W_0$ are fine-tuned and original weights, respectively, while $A \in \mathbb{R}^{m \times r}$ and $B \in \mathbb{R}^{r \times n}$ are low-rank matrices with rank $r \ll \operatorname{min}(m,n)$. LoRA can be used in conjunction with DB for efficient SD fine-tuning. \\

\noindent \textbf{Adversarial Attack.}   In attacks against classifiers, white-box methods like I-FGSM \cite{kurakin2016adversarial,kurakin2018adversarial} or PGD \cite{madry2017towards} are commonly used, which can be formalized as:
\begin{equation}
x_{t+1}^{adv} = \Pi_{x_0, \eta} \left( x_t^{adv} + \alpha \cdot \operatorname{sgn}\left( \nabla_{x_t^{adv}} \mathcal{L}( x_t^{adv}, y; \theta) \right) \right),
\label{eq:PGD}
\end{equation}
where $\Pi_{x_0, \eta}(\cdot)$ restricts inputs within the $l_\infty$-ball of radius $\eta$ around $x_0$, $\operatorname{sgn}(\cdot)$ represents the sign function, and $\alpha$ is the learning rate. $\mathcal{L}( x_t^{adv}, y; \theta)$ is the loss used by the classifier with parameters $\theta$, where $x_t^{adv}$ is the adversarial sample at the $t$-th PGD step and $y$ is the corresponding ground truth label. 
In brief, PGD iteratively finds the most adversarial noises for the model with parameters $\theta$ by maximizing the loss via gradient ascent. \\

\noindent \textbf{Anti-customization.} We explain the intuition behind the simplification that transforms our original maximizing objective from $\underset{\theta_c} {\min} \, \mathbb{E}_{x} \mathcal{L}_{ldm}(x_0 + \delta; \theta_c)$ to $ \mathcal{L}_{ldm}(x_0 + \delta; \theta_c)$ as mentioned above. The key lies in the relationship between the model's training data $x$ (of $\mathbb{E}_{x}$) and the adversarial data $x_0 + \delta^{adv}$.  For optimal performance, the training set should encompass adequately trained adversarial samples. However, this creates a bootstrap paradox: fine-tuned $\theta_c$ is needed for optimal $\delta^{adv}$ while $\delta^{adv}$ is needed for optimal $\theta_c$, which is why surrogate models fine-tuned on clean data are frequently employed for simplification. 

In the context of fine-tuning methods like Textual Inversion \cite{gal2022image}, which make no change to the internal parameters of SD, such an issue exists no more in practice. For DB (full fine-tuning) and LoRA (PEFT), simply using a model fine-tuned on clean images as a surrogate leads to Fully-trained Surrogate Model Guidance (FSMG) formalized in Anti-DB \cite{van2023anti}. A more promising alternative, also proposed by Anti-DB, is to iteratively introduce insufficient adversarial samples, generated at different PGD steps, into the surrogate model alongside clean images. This approach is referred to as Alternating Surrogate and Perturbation Learning (ASPL).\\

\noindent \textbf{Purification.} 
In its original paper, DiffPure \cite{nie2022diffusion} is introduced via Stochastic Differential Equation (SDE). Since we use the specialized DDPM-based purification model, and considering that SD is commonly implemented discretely, the introduction of diffusion-based purification in this paper is also written in the DDPM form. We present the more generalized SDE form from the original DiffPure here for both quick reference and rigor. For a Variance Preserving SDE (VP-SDE) where drift and diffusion coefficient are respectively defined as $f(x,t) := - \frac{\beta(t)}{2}x$ and $g(t) := \sqrt{\beta(t)}$, we first diffuse adversarial $x^{adv}$ with a fixed timestep $t^p \in [0,1]$ via:
\begin{equation}
x(t^{p}) = \sqrt{\alpha(t^{p})} x^{adv} + \sqrt{1- \alpha(t^{p})} \epsilon, 
\label{eq:sde_for}
\end{equation}
where $\alpha(t) := e^{- \int_0^t \beta(s)ds}$, then we solve the reverse-time SDE to get the purified sample with an SDE solver $\operatorname{sdeint}$:
\begin{equation}
\operatorname{Pure}(x^{adv})  = \operatorname{sdeint}(x(t^{p}), f_{\text{rev}}, g_{\text{rev}}, \overline{w}, t^{p}, 0; \theta_p), 
\label{eq:sde_back}
\end{equation}
where $\operatorname{sdeint}$ takes in six inputs: initial value $x(t^{p})$, drift coefficient $f_{\text{rev}}(x,t) := -\frac{\beta(t)}{2}[x + 2s_{\theta_p}(x, t)]$, diffusion coefficient $g_{\text{rev}}(t) := \sqrt{\beta(t)}$, Wiener process $ \overline{w}$, initial time $t^p$, and end time $0$. In the discrete case, this whole purification process corresponds to the specialized DDPMs.

\section{Details on Analysis}
\subsection{More Explanation on Overall Formulation}
Due to the deepening of the computational graph during iterative purification denoising, full-gradient adaptive attacks lead to $\mathcal{O}(N)$ memory cost and may cause vanishing/exploding gradients. 
For a 2GB 256$\times$256 unconditional DDPM purification model, fully tracking its training loss after only 5 consecutive denoising samplings requires up to 25GB memory overhead. 
For differentiability, DiffPure proposes the \textit{adjoint method} to calculate full gradients of the reverse SDE with $\mathcal{O}(1)$ memory cost. However, this method of solving the augmented SDE does not reduce the time complexity.
Backward Path Differentiable Approximation (BPDA) \cite{athalye2018obfuscated}  is also a common approach, but the truly effective surrogate is hard to find. 

Is it entirely infeasible to use full-gradient adaptive attacks? \cite{zhao2024can} mentions such a method for anti-customization, where DDIM \cite{song2020denoising} sampling strategy is utilized to ensure memory usage remains within an acceptable range. However, they report that this adaptive attack is not effective. To demonstrate the instability of purification diffusion models as probabilistic models, we set \( \alpha = 0.005 \), \( \eta = \frac{16}{255} \), and perform a 100-step PGD attack on $\mathcal{L}_{ddpm}$. The resolution of the input images is 256$\times$256. Subsequently, both the clean and the adversarial sample obtained from the attack are purified using DiffPure with \( t^p = 50 \) and \( t^p = 100 \), generating four sets of images, each containing 100 samples. The distributions of these sets are visualized using t-SNE \cite{van2008visualizing} with perplexity set to 10, and the results are presented in Fig. \textcolor{iccvblue}{2}. The convergence of the purified clean and adversarial samples motivates us to turn to the alternative by Eq. (\textcolor{iccvblue}{6}).

\subsection{Experimental Details on the Reason Analysis} \label{sec:analysis_supp}
\begin{figure*}[t]
\centering
\includegraphics[width=0.9\linewidth]{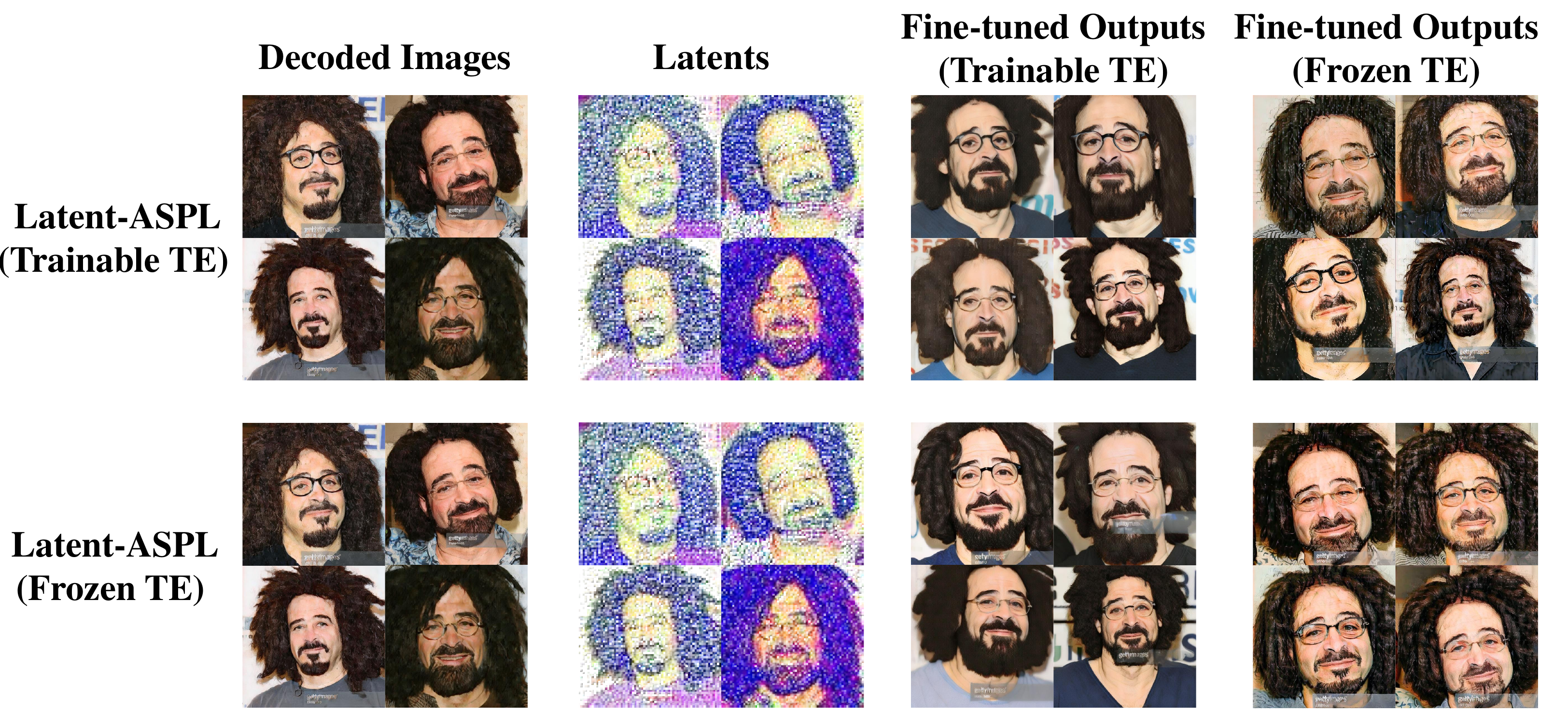}
\caption{ASPL attacks \cite{van2023anti} against SD in the latent space. In the ASPL attack, two configurations are used: trainable/frozen text encoder, corresponding to the two rows in the figure. Similarly, in the DreamBooth fine-tuning, the trainable/frozen text encoder configurations are also employed, corresponding to the last two columns.} 
\vspace{-10pt}
\label{fig:app_zatt}
\end{figure*}

\textbf{Reason 1:  Lack of Vulnerable Components.} 
Firstly, we modify Anti-DB's ASPL method to conduct PGD attacks directly in the latent space. We provide a more comprehensive experimental result here in \cref{fig:app_zatt}, with the CLIP text encoder \cite{radford2021learning} taken into consideration.
We set \( \alpha = 0.005 \), \( \eta_z = \frac{16}{255} \), and perform a (20$\times$5)-step PGD attack on $\mathcal{L}_{ldm}$. Fine-tuning steps per 5 PGD steps are set to 3. In the ASPL attack, we employ two configurations: one with a trainable text encoder (Latent-ASPL, trainable text encoder) and one with a frozen text encoder (Latent-ASPL, frozen text encoder), and the adversarial examples shown in Fig. \textcolor{iccvblue}{3} are obtained via the former. Actually, these two different configurations do not result in significant differences, whether in the generated adversarial samples or in the outputs obtained after fine-tuning SD on the adversarial samples. 

To avoid introducing additional noise during VAE decoding and to maintain consistency in the number of channels, we directly save the adversarial latents in ``.pt" format and use them to replace the corresponding instance inputs in the DreamBooth training process.

During customization, we fine-tune SD v2.1 via DreamBooth on these two kinds of adversarial samples. We also choose to either train or freeze the text encoder during fine-tuning. When jointly training the text encoder, we set the learning rate to $5e-7$, and when freezing the text encoder, we set it to $5e-6$ to ensure the capture of the target concept, with 500 steps of training for both fine-tuning configurations. The training instance prompt is ``a photo of \textit{sks} person", the class prompt is ``a photo of person", and the inference prompt is ``a photo of \textit{sks} person". The training batch size is 2, with a prior loss weight of $1.0$. The results are shown in \cref{fig:app_zatt}.

Also, we directly attack $\mathcal{L}_{ddpm}$ using 200 randomly selected images in our datasets, resized to 256$\times$256. We set \( \alpha = 0.005 \), \( \eta = \frac{16}{255} \), and perform a 150-step PGD attack, with Monte-Carlo sampled timesteps limited in [1, 100]. Subsequently, the adversarial images and clean images are both fed into the UNet with time condition inputs ranging from 1 to 100. The purification model we use consists of 18 downsampling blocks, 1 middle block, and 18 upsampling blocks. We record MSE between the intermediate outputs of adversarial and clean images block by block under different time conditions. The average values across 200 images are computed, and the final results are presented in Fig. \textcolor{iccvblue}{4}. It can be observed that, due to the increasing coefficient $\sqrt{1-\overline{\alpha}_t}$, the differences between clean and adversarial images grow with longer timesteps. \\

\noindent \textbf{Reason 2: Frozen Parameters with Benign Priors.} 
The adversarial samples in Fig. \textcolor{iccvblue}{5} are generated using the $\mathcal{L}_{ldm}$-attack against SD v1.5. The configuration for generating protective perturbation is largely consistent with the setup used in the experiments above, conducted in the latent space. We set \( \alpha = 0.005 \), \( \eta = \frac{16}{255} \), and perform a 100-step PGD attack. During editing, we employ MasaCtrl \cite{cao2023masactrl} combined with a pretrained T2I-Adapter \cite{mou2024t2i}, with the condition type set to ``sketch." The significant attenuation of artifacts demonstrates that this direct attack on training objectives is not fully applicable to training-free tasks. \\

\begin{figure}[t]
\centering
\includegraphics[width=1.0\linewidth]{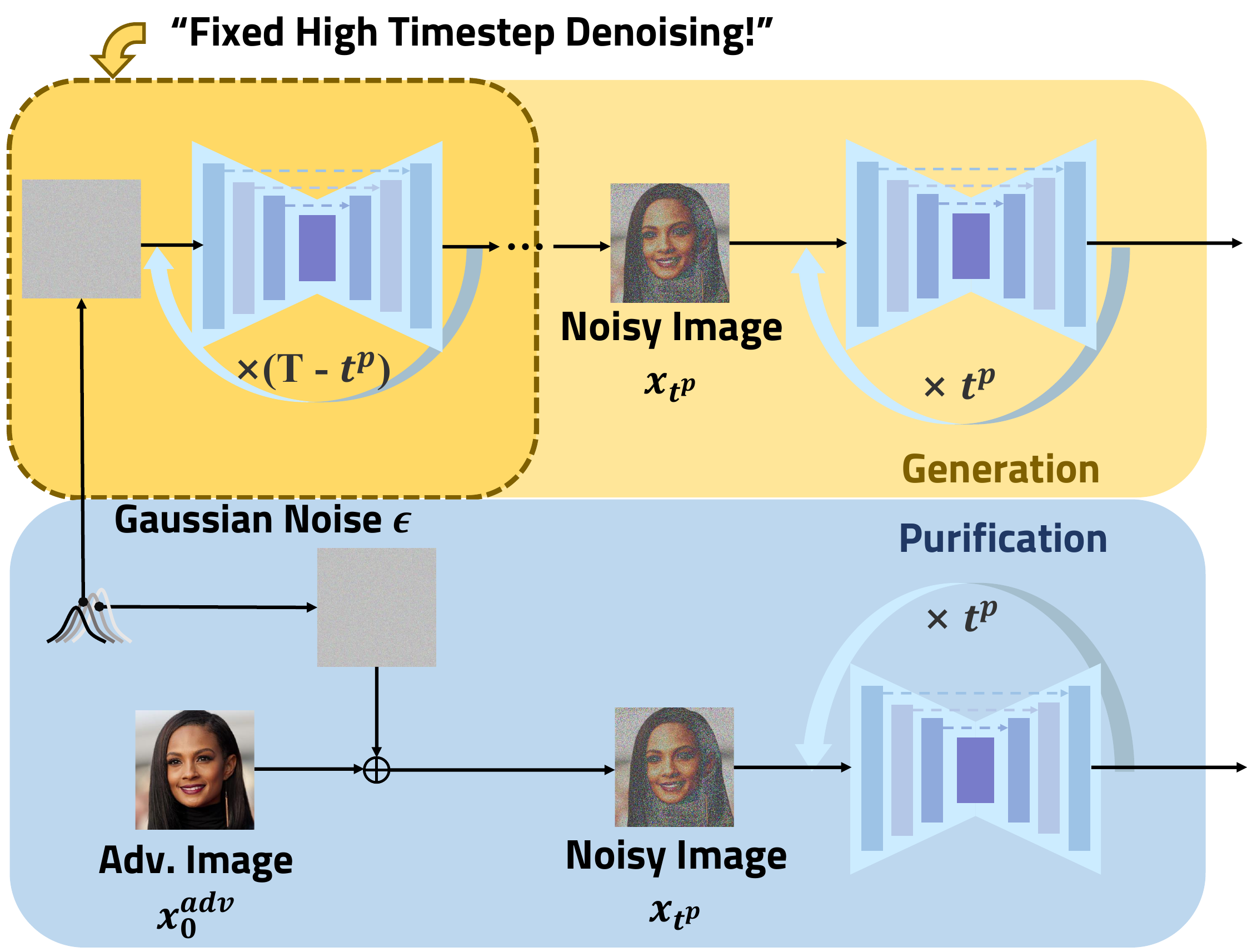}
\caption{Why ``the purification process can be viewed as a generation process where high-timestep denoising is fixed."} 
\label{fig:pure}
\vspace{-10pt}
\end{figure}

\noindent \textbf{Reason 3: Fixed High Timestep Denoising.}
Perhaps the statement in Sec. \textcolor{iccvblue}{4.2.3}, ``the purification process can be viewed as a generation process where high-timestep denoising is fixed," is not sufficiently direct. To offer a more intuitive illustration of this process, we present a simple diagram. As depicted in \cref{fig:pure}, attacks on generation span the entire range from $T$ (typically set to 1000) to 0. In contrast, attacks on purification are limited to a much smaller range, from $t^p$ to 0, where the low-frequency structural information fixed during the ``Fixed Higher Timestep Denoising" stage cannot be effectively altered.

\section{Experimental Details}\label{sec:detail_supp}
Here, we provide more configuration details used for the quantitative experimental evaluation.
In Sec. \textcolor{iccvblue}{6.2} and Sec. \textcolor{iccvblue}{6.3}, we use the same setup for all perturbation methods. In PGD attacks, we normalize images to \([-1, 1]\). Within this range, the noise budget $\eta$ is set to \(16/255\), the learning rate \(\alpha\) is set to $5e^{-3}$, and the total PGD steps are set to 100 (20 $\times$ 5 steps for Anti-DB and SimAC). In Sec. \textcolor{iccvblue}{6.5}, we evaluate the perceptual consistency of different perturbation methods with pretrained AlexNet\cite{krizhevsky2012imagenet}/VGG\cite{simonyan2014very}. 

The purification and fine-tuning settings are also kept consistent. In Sec. \textcolor{iccvblue}{6.2} and Sec. \textcolor{iccvblue}{6.3}, we use GrIDPure \cite{zhao2024can} for purification, applying 2 rounds of 20 iterations with $t^{p} = 10$, $\gamma=0.1$ to approximate convergence of the purification effect. 
In Sec. \textcolor{iccvblue}{6.4}, we use a finer-grained purification configuration to explore when the purification effect approximately reaches convergence. Specifically, we apply 4 rounds of 20 iterations with $t^{p} = 10$, $\gamma=0.1$, on the adversarial images obtained from Anti-DB and AntiPure. Totally, Tab. \textcolor{iccvblue}{3} uses 4 rounds × 10 iters, where \textit{P(Iter=30)-C} equals 3×10, and so on. GrIDPure mitigates image degradation caused by purification via residual connections, allowing 4×10 to rely more on intermediate results while maintaining the same computational cost as 2×20. This leads to inconsistency between the results in Tab. \textcolor{iccvblue}{1} and the results of P(Iter=40)-C in Tab. \textcolor{iccvblue}{3}. But overall, the computational overhead incurred by these two settings during the purification process is the same. Here, 20-iter is the default setting of GrIDPure.

For customization, in Sec. \textcolor{iccvblue}{6.2}, we fine-tune the UNet and the text encoder jointly by DreamBooth \cite{ruiz2023dreambooth} with batch size of 2 and learning rate of $5e^{-7}$ for 500 training steps. The training instance prompt is ``a photo of \textit{sks} person", the class prompt is ``a photo of person", and the inference prompt is ``a photo of \textit{sks} person". We also set the prior loss weight to $1.0$. 
In Sec. \textcolor{iccvblue}{6.3}, we apply LoRA \cite{hu2021lora} with the same DreamBooth settings but set the learning rate to $5e^{-5}$. The rank is set to 4. 
For evaluation, 30 PNG images per ID are generated, which is also consistent with the configurations used in Anti-DB and SimAC.

\section{More Experimental Results}\label{sec:result_supp}

\subsection{Ablation Study}
Our proposed AntiPure incorporates two kinds of additional guidance to address the inherent challenges of the anti-purification task: 1) Patch-wise Frequency Guidance and 2) Erroneous Timestep Guidance. In the ablation study, we gradually remove this guidance to validate the effectiveness of our method. 

We use the same attack/purification/customization experimental configurations in Sec. \textcolor{iccvblue}{6.2} and Sec. \textcolor{iccvblue}{6.3} to perform the corresponding DreamBooth and LoRA fine-tuning on CelebA-HQ and VGGFace2, but with different attack targets. Specifically, our attack targets include: 1) $\mathcal{L}_{ddpm}$, 2) $\mathcal{L}_{ddpm} + \mathcal{L}_{fre}$, 3) $\mathcal{L}_{ddpm} + \mathcal{L}_{err-t}$, and we compare these results with the full AntiPure, i.e., 4) $\mathcal{L}_{ddpm} + \mathcal{L}_{fre} + \mathcal{L}_{err-t}$. The DreamBooth fine-tuning results are shown in \cref{tab:abla_db}, and the LoRA results are shown in \cref{tab:abla_lora}.
\begin{table}[t]
\renewcommand{\arraystretch}{1.5}
\centering
\resizebox{\columnwidth}{!}{
\begin{tabular}{ccccc}
\hline
Dataset & Objective & FID↑ & ISM↓ (FDFR) & BRISQUE↑ \\ \hline

\multirow{4}{*}{CelebA-HQ}& $\mathcal{L}_{ddpm}$ &  69.06&  0.6293 (0.09)& 42.45\\  
& $\mathcal{L}_{ddpm} + \mathcal{L}_{fre}$ &  65.69&  0.6253 (0.08)& 42.84 \\ 
& $\mathcal{L}_{ddpm} + \mathcal{L}_{err-t}$ &   74.42&  0.6489 (0.10)&  37.01\\ \cline{2-5} 
& \textbf{AntiPure (Ours)}& \textbf{81.15} & \textbf{0.6112} (0.10) & \textbf{43.60}\\ \hline

\multirow{4}{*}{VGGFace2}& $\mathcal{L}_{ddpm}$ &  76.32&  0.5958 (0.07)& 39.42\\ 
& $\mathcal{L}_{ddpm} + \mathcal{L}_{fre}$ &  74.90& 0.5644 (0.07)& 45.57\\ 
& $\mathcal{L}_{ddpm} + \mathcal{L}_{err-t}$ & 76.75 & 0.5901 (0.06) & 40.75 \\ \cline{2-5} 
& \textbf{AntiPure (Ours)}& \textbf{90.77}& \textbf{0.5475} (0.05) & \textbf{46.01}\\ \hline
\end{tabular}
}
\caption{Ablation Study on DreamBooth's \cite{ruiz2023dreambooth} output quality for different AntiPure guidance following the Purification-Customization (P-C) workflow.}
\label{tab:abla_db}
\vspace{-7pt}
\end{table}

It is evident that AntiPure, which combines both types of guidance, achieves the best overall performance across various metrics, datasets, and fine-tuning methods, resulting in the most significant output distortion. This represents a clear improvement over the original $\mathcal{L}_{ddpm}$-based attack. Additionally, it can be observed that among the single guidance methods, $\mathcal{L}_{fre}$ is more effective than $\mathcal{L}_{err-t}$. In fact, using $\mathcal{L}_{err-t}$ for extra guidance alone shows limited impact. However, it helps confuse the model across different time steps, thereby disrupting the frequency characteristics of the predicted noise, providing a better foundation for $\mathcal{L}_{fre}$ guidance. This is particularly evident in FID, where AntiPure sees an obvious improvement when both types of guidance are combined. 

In other words, the combination of these two guidance types is not merely an additive process but achieves a synergistic ``$1+1>2$" effect. Actually, the timestep inputs of the diffusion model's UNet can affect the frequency representation of the predicted noise, allowing $\mathcal{L}_{err-t}$ to be interpreted on the frequency domain like $\mathcal{L}_{fre}$. With both involved, the high-frequency components intensified by $\mathcal{L}_{fre}$ are primarily induced by erroneous high timesteps rather than real ones. Thus, the introduction of $\mathcal{L}_{err-t}$ can indirectly enhance $\mathcal{L}_{fre}$ itself, and vice versa.

\begin{table}[t]
\renewcommand{\arraystretch}{1.5}
\centering
\resizebox{\columnwidth}{!}{
\begin{tabular}{ccccc}
\hline
Dataset & Objective & FID↑ & ISM↓ (FDFR) & BRISQUE↑ \\ \hline

\multirow{4}{*}{CelebA-HQ}& $\mathcal{L}_{ddpm}$&  93.79&  0.6176 (0.05)& 42.19\\  
& $\mathcal{L}_{ddpm} + \mathcal{L}_{fre}$ &  81.32&  0.5848 (0.05)& 42.24\\ 
& $\mathcal{L}_{ddpm} + \mathcal{L}_{err-t}$ &   92.63&  0.6177 (0.09)&  \textbf{43.22}\\ \cline{2-5} 
& \textbf{AntiPure (Ours)} & \textbf{109.63}& \textbf{0.5839} (0.07)& 40.01\\ \hline

\multirow{4}{*}{VGGFace2}& $\mathcal{L}_{ddpm}$ &  93.10&  0.5859 (0.08)& 61.79\\ 
& $\mathcal{L}_{ddpm} + \mathcal{L}_{fre}$ &  110.87& 0.5556 (0.06)& 66.01\\ 
& $\mathcal{L}_{ddpm} + \mathcal{L}_{err-t}$ &   102.24& 0.5717 (0.06)& 61.10\\ \cline{2-5} 
& \textbf{AntiPure (Ours)} & \textbf{127.67}& \textbf{0.5428} (0.04)& \textbf{69.97}\\ \hline
\end{tabular}
}
\caption{Ablation Study on LoRA's \cite{hu2021lora} output quality for different AntiPure guidance following the Purification-Customization (P-C) workflow.}
\label{tab:abla_lora}
\end{table}

\subsection{Transformation Robustness}
As suggested by the reviewer, we apply \textit{Crop-Scale} (CenterCrop 
$\times 3/4$ side length) and \textit{Rotation} (randomly $[0^{\circ}, 15^{\circ}]$) to anti-purification samples created by AntiPure, ensuring that the same transformations are applied to the original ones for fair evaluation. As shown in \cref{tab:re_trans}, AntiPure demonstrates robustness to \textit{rotation}, while \textit{crop-scale} amplifies the artifacts, leading to significantly improved performance.

\begin{table}[t]
\renewcommand{\arraystretch}{1.5}
\centering
\resizebox{\columnwidth}{!}{
\begin{tabular}{ccccc}
\hline
Dataset & Transformation & FID↑ & ISM↓ (FDFR) & BRISQUE↑ \\ \hline
\multirow{3}{*}{VGGFace2}
& Crop-Scale & 152.47 & 0.4805 (0.34) & 53.60  \\ 
& Rotation & 92.00 & 0.5550 (0.05) & 45.14  \\ \cline{2-5} 
& \textbf{None (Ours)} & \textbf{90.77}& \textbf{0.5475} (0.05) & \textbf{46.01}\\ \hline
\end{tabular}
}
\caption{Comparison of DreamBooth's \cite{ruiz2023dreambooth} output quality on VGGFace2 for different transformations on AntiPure's outputs.}
\label{tab:re_trans}
\end{table}

\subsection{More Baselines}
As suggested by the reviewer, we include PhotoGuard \cite{salman2023raising} and CAAT \cite{xu2024perturbing} for additional comparison. We adopt the \textit{img2img} attack pipeline for PhotoGuard, as it resembles purification more than the \textit{inpainting} pipeline. However, as shown in \cref{tab:re_moreb}, PhotoGuard's perturbations tend to be easily purified due to their blurred boundaries. In contrast, CAAT's perturbation closely resembles that of Anti-DB, leading to comparable robust performance.

\begin{table}[t]
\renewcommand{\arraystretch}{1.5}
\centering
\resizebox{\columnwidth}{!}{
\begin{tabular}{ccccc}
\hline
Dataset & Perturbation & FID↑ & ISM↓ (FDFR) & BRISQUE↑ \\ \hline
\multirow{3}{*}{VGGFace2}
& PhotoGuard \cite{salman2023raising} & 72.25 & 0.6061 (0.07) & 43.07  \\ 
& CAAT \cite{xu2024perturbing} & 89.07 & 0.5854 (0.07) & 38.21  \\ \cline{2-5} 
& \textbf{AntiPure (Ours)}& \textbf{90.77}& \textbf{0.5475} (0.05) & \textbf{46.01}\\ \hline
\end{tabular}
}
\caption{Comparison with additional baselines on VGGFace2.}
\label{tab:re_moreb}
\end{table}

\subsection{Hyperparameter Sensitivity}
Originally, the selection of \(\lambda_1\) and \(\lambda_2\) was based on balancing the magnitude of loss components, while \(t_{\text{err}}\) was chosen to be as large as possible to maximize its effect. Here, as the reviewer suggested, we conduct a simple grid search over these three hyperparameters. As shown in \cref{tab:re_hyperm}, different metrics exhibit varying degrees of sensitivity to each parameter. Notably, the impact of \(t_{\text{err}}\) is relatively smaller compared to those of \(\lambda\)s, while the ISM—the primary metric for identity preservation—remains largely stable across all settings. This suggests that AntiPure exhibits a certain degree of robustness with respect to its hyperparameter configurations.

\begin{table}[t]
\renewcommand{\arraystretch}{1.5}
\centering
\resizebox{\columnwidth}{!}{
\begin{tabular}{ccccccc}
\hline
Dataset & $\lambda_1$ & $\lambda_2$ & $t_{err}$ & FID↑ & ISM↓ (FDFR) & BRISQUE↑ \\ \hline
\multirow{4}{*}{VGGFace2}
& 0.25 & 0.75 & 999 & 96.33 & 0.5431 (0.06) & 41.50 \\ 
& 0.50 & 0.50 & 700 & 90.50 & 0.5490 (0.04) & 48.34 \\
& 0.75 & 0.25 & 999 & 87.81 & 0.5586 (0.05) & 43.02 \\ \cline{2-7} 
& 0.50 & 0.50 & 999 & \textbf{90.77} & \textbf{0.5475} (0.05) & \textbf{46.01}\\ \hline
\end{tabular}
}
\caption{DreamBooth's \cite{ruiz2023dreambooth} output quality on VGGFace2 for different hyperparameter configurations.}
\label{tab:re_hyperm}
\end{table}

\subsection{Black-Box Performance}

All previous experiments are conducted on SD v2.1, as recommended by Anti-DB and SimAC. However, AdvDM and Mist only support SD v1.x. We note that after sufficient purification, the effects of these perturbation methods almost completely disappear, making the distinction between SD versions insignificant. 

To evaluate the performance of perturbation methods under a black-box scenario with mismatched models, and to ensure an absolutely fair SD version for all methods, we fine-tune SD v1.5 on the purified adversarial images from VGGFace2. The results are shown in \cref{tab:sd15}. The similar performance observed preliminarily supports our hypothesis that ``SD versions have negligible influence." Also, AntiPure still demonstrates the best overall performance.

\begin{table}[t]
\renewcommand{\arraystretch}{1.5}
\centering
\resizebox{\columnwidth}{!}{
\begin{tabular}{ccccc}
\hline
Fine-tuning& Perturbation & FID↑ & ISM↓ (FDFR) & BRISQUE↑ \\ \hline

\multirow{5}{*}{DreamBooth}& AdvDM \cite{liang2023adversarial}& 82.10& 0.5798 (0.06)& 26.99\\ 
& Mist \cite{liang2023mist}&  77.33&  0.5797 (0.04)&  32.58\\ 
& Anti-DB \cite{van2023anti}& 83.95& 0.5686 (0.06)& 27.68\\ 
& SimAC \cite{wang2024simac}& 76.73& 0.5762 (0.05)& 26.44\\ \cline{2-5} 
& \textbf{AntiPure (Ours)}& \textbf{89.33}& \textbf{0.5165} (0.03)& \textbf{62.88}\\ \hline

\multirow{5}{*}{LoRA}& AdvDM \cite{liang2023adversarial}& 106.48& 0.5697 (0.05)& 44.96\\ 
& Mist \cite{liang2023mist}&  91.33&  0.5731 (0.06)&  55.27\\ 
& Anti-DB \cite{van2023anti}& \textbf{115.34}& 0.5591 (0.05)& 46.11\\ 
& SimAC \cite{wang2024simac}& 92.57& 0.5622 (0.05)& 45.41\\ \cline{2-5} 
& \textbf{AntiPure (Ours)}& 112.90& \textbf{0.5101} (0.05)& \textbf{74.82}\\ \hline
\end{tabular}
}
\caption{Comparison of DreamBooth/LoRA's \cite{ruiz2023dreambooth} Stable Diffusion v1.5 output quality on VGGFace2 for different perturbation methods following the P-C workflow.}
\label{tab:sd15}
\end{table}

\subsection{More visualization}

\begin{figure*}[t]
\centering
\includegraphics[width=0.9\linewidth]{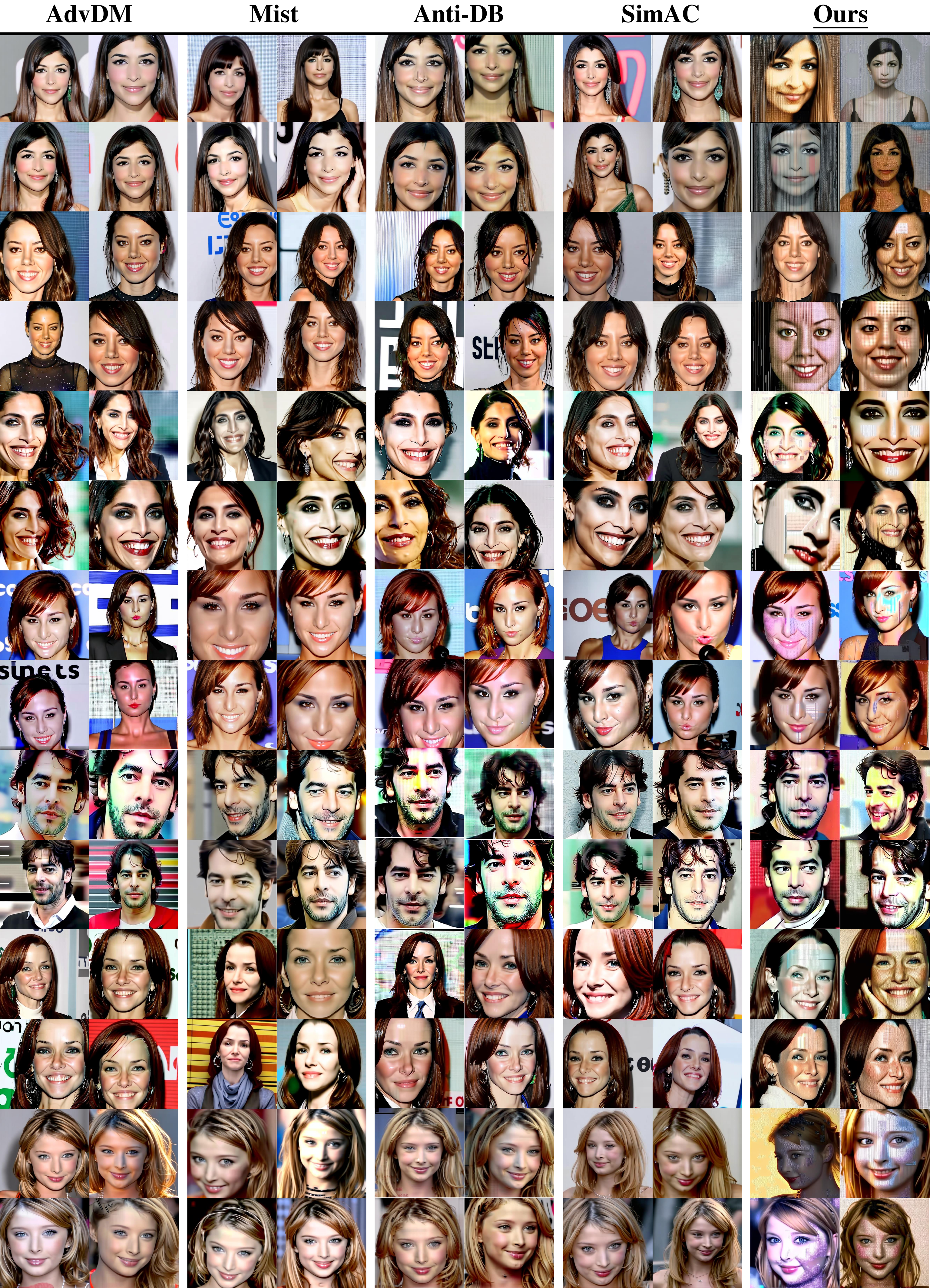}
\caption{Comparison of DreamBooth's outputs on CelebA-HQ for different perturbation methods following the Purification-Customization (P-C) workflow.} 
\label{fig:more1}
\end{figure*}

\begin{figure*}[t]
\centering
\includegraphics[width=0.9\linewidth]{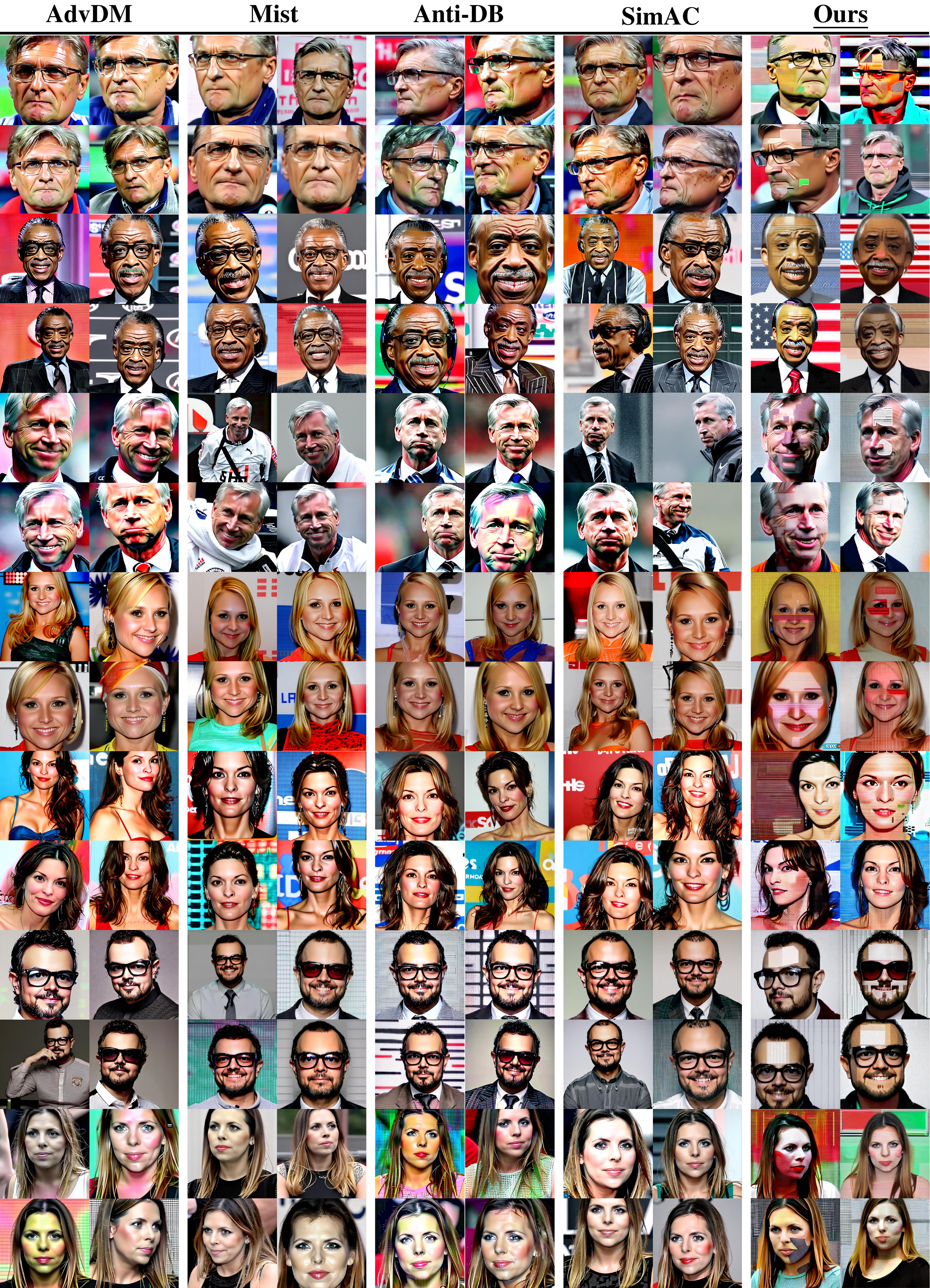}
\caption{Comparison of DreamBooth's outputs on VGGFace2 for different perturbation methods following the Purification-Customization (P-C) workflow.} 
\label{fig:more2}
\end{figure*}

\begin{figure*}[t]
\centering
\includegraphics[width=0.9\linewidth]{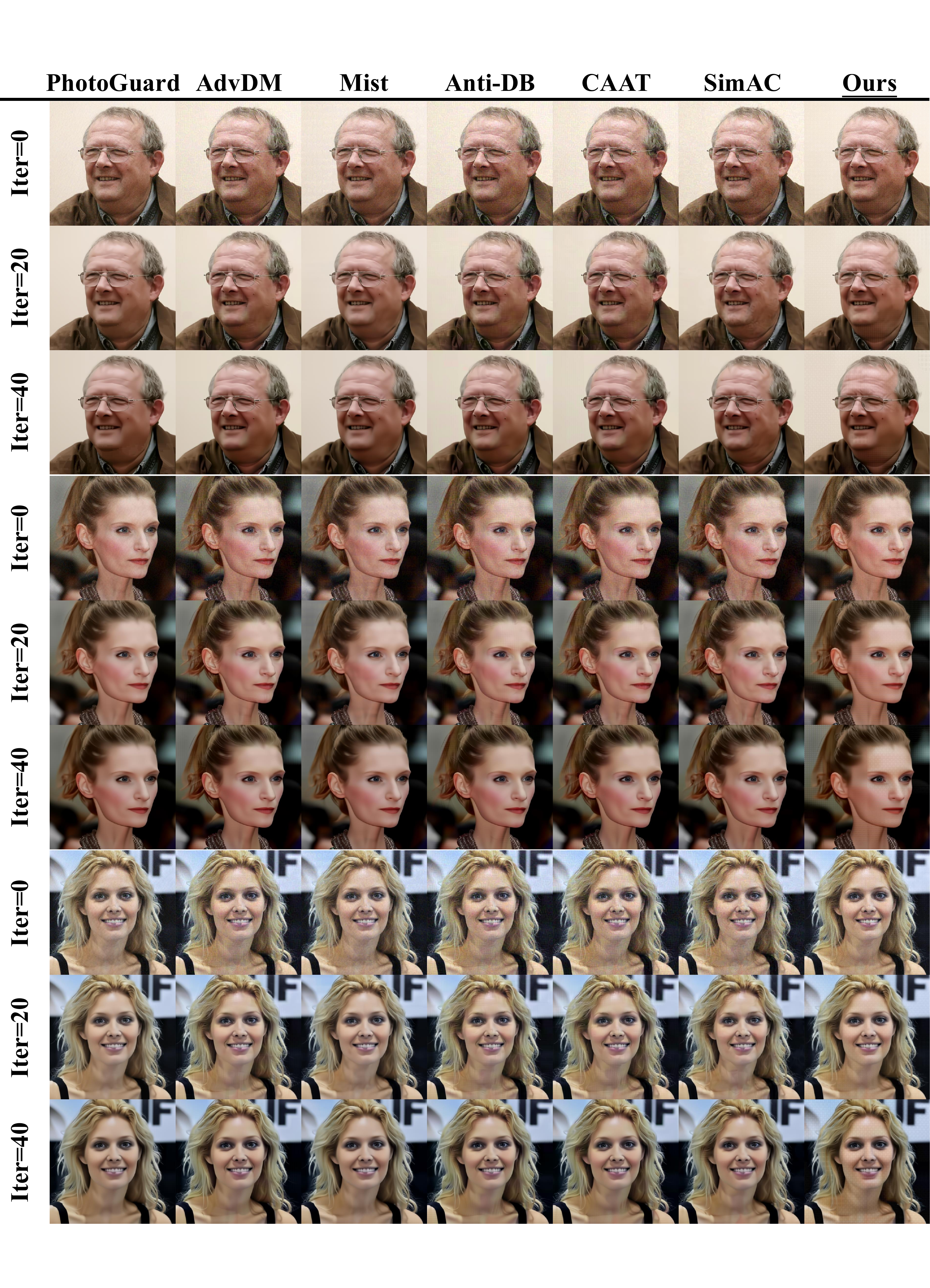}
\caption{Comparison of GrIDPure's outputs at different iterations on VGGFace2 for different perturbation methods. Here \textit{Iter=0} means no purification is adopted after adversarial samples are generated.} 
\label{fig:morep1}
\end{figure*}

\begin{figure*}[t]
\centering
\includegraphics[width=0.9\linewidth]{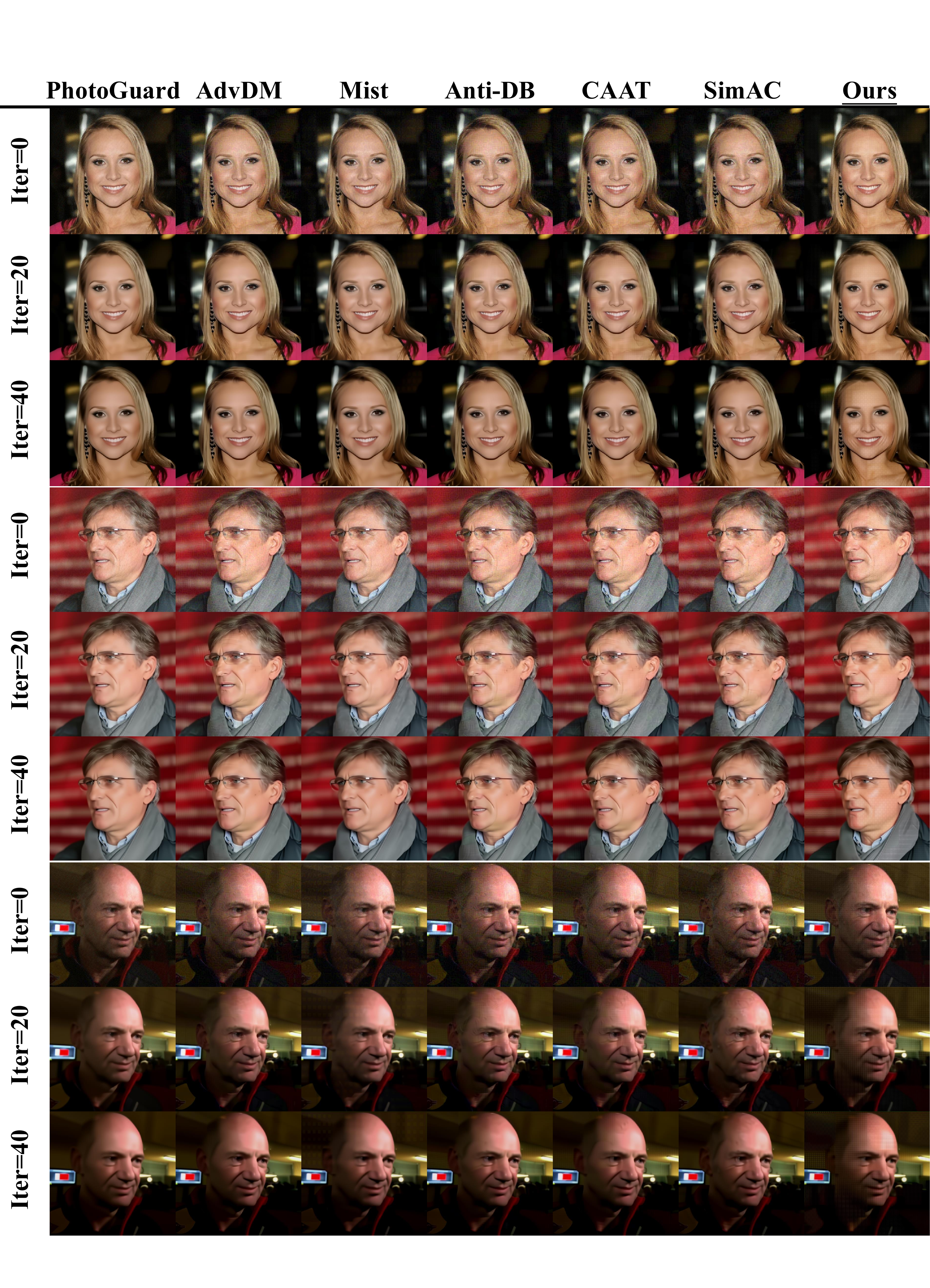}
\caption{Comparison of GrIDPure's outputs at different iterations on VGGFace2 for different perturbation methods. Here \textit{Iter=0} means no purification is adopted after adversarial samples are generated.} 
\label{fig:morep2}
\end{figure*}

We provide more visualization results in \cref{fig:more1,fig:more2,fig:morep1,fig:morep2} for qualitative evaluation. Please refer to the captions of each figure for detailed explanations. \textbf{We strongly recommend zooming in} on the following visualizations to better identify these artifacts.

It can be observed that the effects of other protective perturbation methods almost entirely vanish after sufficient purification. However, AntiPure ensures the presence of detectable artifacts, which are concentrated in the human facial regions (excluding the eyes). At lower levels of semantic distortion, these artifacts appear as unnatural high-frequency speckled regions, while more prominent artifacts manifest as patches of abnormal textures.  

Furthermore, the effects of different perturbation methods on human visual perception (\textit{iter = 0}, i.e., no purification) in \cref{fig:morep1,fig:morep2} are also consistent with the LPIPS comparison in Tab. \textcolor{iccvblue}{4}.
Even under the same noise budget, Anti-DB and CAAT perturbations are more noticeable, often exhibiting blocky color artifacts. AntiPure, however, relies on frequency-domain modulation and generates samples visually closer to the original image.

\clearpage 

\end{document}